\documentclass{article}

\usepackage{arxiv}

\usepackage[utf8]{inputenc} 
\usepackage[T1]{fontenc}    
\usepackage{hyperref}       
\usepackage{url}            
\usepackage{booktabs}       
\usepackage{amsfonts}       
\usepackage{nicefrac}       
\usepackage{microtype}      
\usepackage{lipsum}		
\usepackage{graphicx}
\usepackage{natbib}
\usepackage{doi}

\title{Electronic Health Records:\\Towards Digital Twins in Healthcare}

\author{ {Muhammet Alkan} \\
	School of Computing Science\\
	University of Glasgow\\
	Glasgow, Scotland, UK \\
    \And
	{Hester Huijsdens} \\
	School of Computing Science\\
	University of Glasgow\\
	Glasgow, Scotland, UK \\
    \And
	{Yola Jones} \\
	School of Computing Science\\
	University of Glasgow\\
	Glasgow, Scotland, UK \\
    \And
	{Fani Deligianni}\thanks{Corresponding author, \texttt{Fani Deligianni}} \\
	School of Computing Science\\
	University of Glasgow\\
	Glasgow, Scotland, UK \\
    \texttt{Fani.Deligianni@glasgow.ac.uk}
}

\renewcommand{\shorttitle}{Electronic Health Records: Towards Digital Twins in Healthcare}

\begin{document}
\maketitle

\begin{quote}
    ``The good physician treats the disease; the great physician treats the patient who has the disease.'' --- William Osler 
\end{quote}

\begin{abstract}
The pivotal shift from traditional paper-based records to sophisticated Electronic Health Records (EHR), enabled systematic collection and analysis of patient data through descriptive statistics, providing insight into patterns and trends across patient populations. This evolution continued toward predictive analytics, allowing healthcare providers to anticipate patient outcomes and potential complications before they occur. This progression from basic digital record-keeping to sophisticated predictive modelling and digital twins reflects healthcare's broader evolution toward more integrated, patient-centred approaches that combine data-driven insights with personalized care delivery.
This chapter explores the evolution and significance of healthcare information systems, beginning with an examination of the implementation of EHR in the UK and the USA. It provides a comprehensive overview of the International Classification of Diseases (ICD) system, tracing its development from ICD-9 to ICD-10. Central to this discussion is the MIMIC-III database, a landmark achievement in healthcare data sharing and arguably the most comprehensive critical care database freely available to researchers worldwide. MIMIC-III has democratized access to high-quality healthcare data, enabling unprecedented opportunities for research and analysis.  
The chapter examines its structure, clinical outcome analysis capabilities, and practical applications through case studies, with a particular focus on mortality and length of stay metrics, vital signs extraction, and ICD coding. 
Through detailed entity-relationship diagrams and practical examples, the text illustrates MIMIC's complex data structure and demonstrates how different querying approaches can lead to subtly different results, emphasizing the critical importance of understanding the database's architecture for accurate data extraction.
The chapter concludes by discussing the progression from descriptive analytics to digital twins in healthcare, highlighting the transformation toward more sophisticated and personalized healthcare information systems.
\end{abstract}

\keywords{EHR \and ICD \and MIMIC \and descriptive statistics \and predictive analytics \and digital twins}

\section{Electronic Health Records}

\subsection{Electronic Health Records}

Electronic Health Records (EHRs) are digital versions of a patient's health information. 
EHRs exhibit a diverse composition, encompassing both structured and unstructured data types, such as medical images, Electrocardiograms (ECGs), laboratory results, narrative, diagnostic reports, medical interventions and hospitalisations.

\begin{figure*}[hbtp]
\centering
\includegraphics[width=1\textwidth]{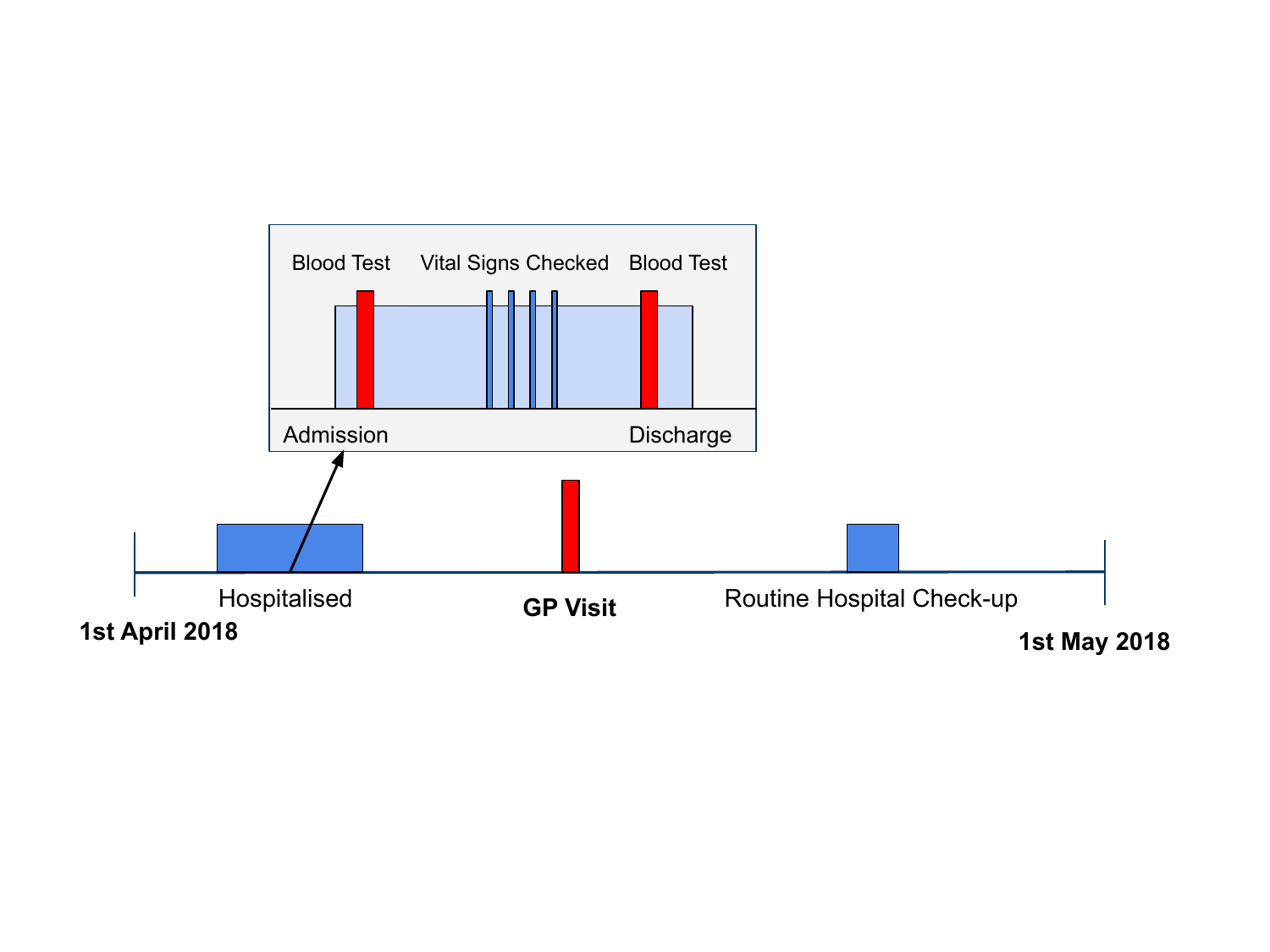}
\vspace{-100pt}
\caption{Patient's Timeline}
\label{fig:patient_timeline_example}
\end{figure*}

Unlike clinical trials, which follow-up patients at regular intervals (e.g., 6 months, 1 year, 5 years), electronic health records capture data at irregular intervals by recording the dates of each event. For example, someone may be hospitalised for three days, receive extensive medical follow-up during these three days, and then not be seen by the medical system for the following year (see Figure \ref{fig:patient_timeline_example} as an example).

\subsection{The transition to Electronic Health Records in the UK and USA}

The shift from paper-based to Electronic Health Records (EHRs) has profoundly influenced healthcare data management, enhanced the quality of clinical decision making and opened new avenues for biomedical research. It also signaled a transition from independent records to integrated databases. 

\begin{figure*}[hbtp]
\centering
\includegraphics[width=\textwidth]{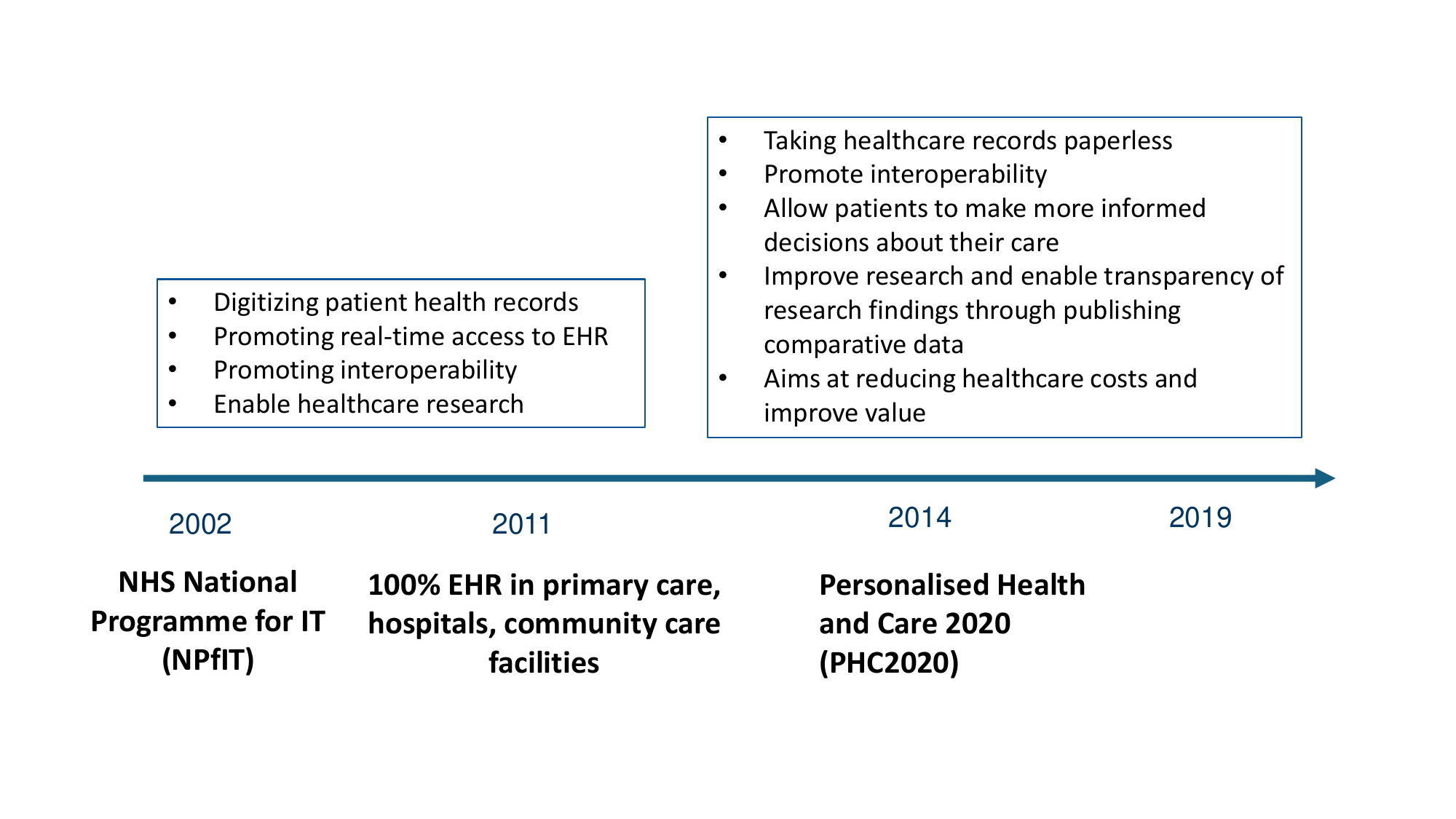}
\vspace{-40pt}  
\caption{The transition to EHRs in UK}
\label{fig:EHR_UK}
\end{figure*}

\begin{figure*}[hbtp]
\centering
\includegraphics[width=\textwidth]{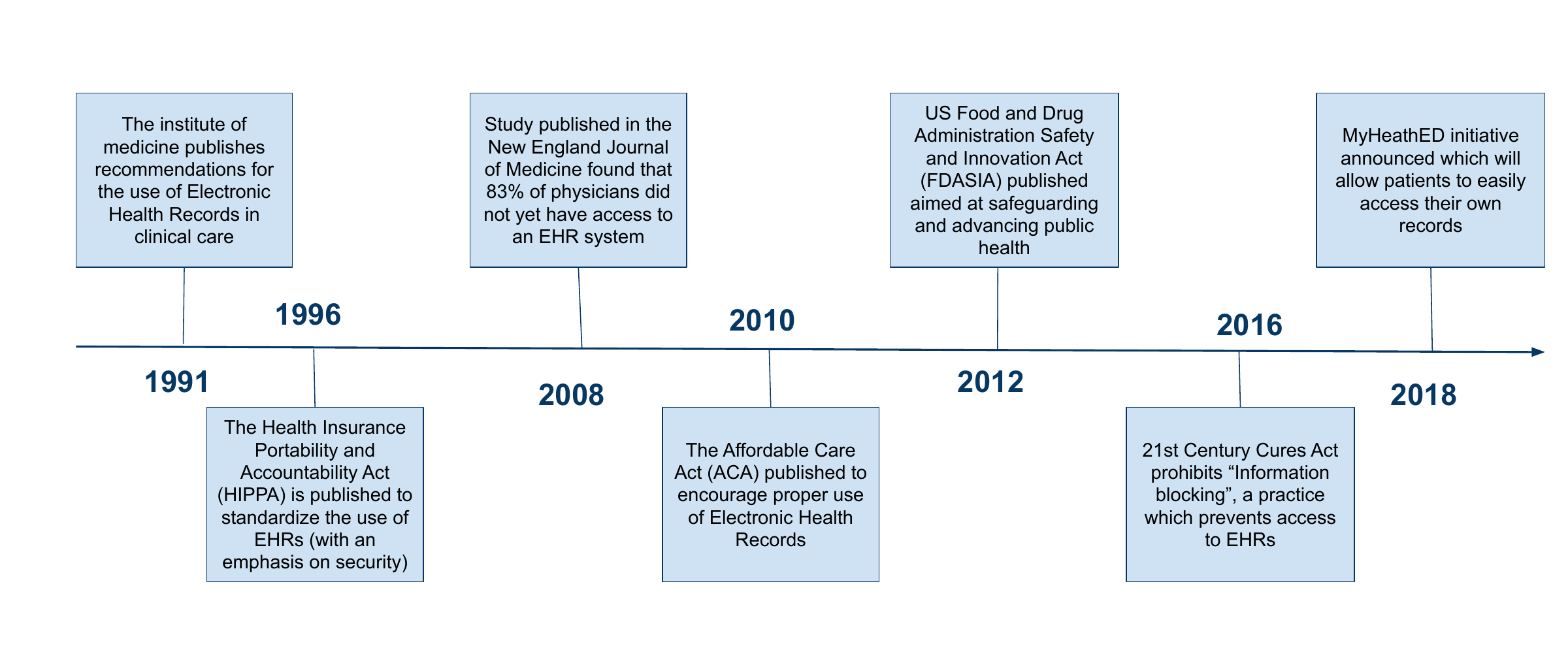}
\caption{The transition to EHRs in USA}
\label{fig:EHR_USA}
\end{figure*}

An overview of the adoption of EHRs in the USA and UK healthcare systems are shown in Figure \ref{fig:EHR_UK} and Figure \ref{fig:EHR_USA}, respectively. This transition took over the period of two decades and it highlights the challenges and requirements for a successful integration. The US and the UK health care systems are organised very differently \citep{wilson2018migrating}. UK has one of the largest public sector system. On the other hand, USA has the largest private-sector system and one of the largest health care expenditure in the world.

Health Insurance Portability and Accountability Act is the first of the USA laws to affect IT in healthcare \citep{ozair2015ethical}. It was enacted in August of 1996. Part of this sets the rules and requirements for the privacy and security of information. This information was referred to as Protected Health Information. Rules were set in place for what and how it could be accessed and shared. The purpose was to encourage the use of electronic data interchange by explaining how data could safely be shared between parties. Part of it was to define standards for corporations to abide by and thus help building interoperability.

 Despite HIPAA, by 2006, over 80\% of US physicians lacked access to EHRs, leading to efforts to establish a nationwide health information network and the extension of Protected Health Information (PHI) responsibility to business associates. The \textbf{Patient Protection and Affordable Care Act} in 2010 and the \textbf{US Food and Drug Administration Safety and Innovation Act} in 2012 further advanced public health, stakeholder engagement, and the safety of the drug supply chain.

On the other hand, the UK initiated the \textbf{National Program for IT} in 2002, aiming for digitization of health records and paperless operations by 2020, but faced challenges in meeting stakeholder expectations.
While substantial advancements have been made, there is still complexity in the healthcare system that requires careful management to optimize the use of EHRs \citep{gamal2021standardized}.

\subsection{Clinical Registries and Databases in Healthcare}
 
Clinical registries play a crucial role in modern healthcare systems by serving as comprehensive repositories of patient data focused on specific conditions or diseases. These sophisticated systems are designed to collect, store, and analyze information over extended periods, offering invaluable insights into patient care and treatment outcomes. The primary functions of clinical registries include monitoring and enhancing the quality of care, evaluating the efficacy of various treatments, and providing a solid foundation for research initiatives in targeted medical areas.

Several notable examples of clinical registries have been established across Europe, each contributing significantly to their respective fields. The Dutch CONCOR Registry, initiated in 2005, stands as the national registry and DNA bank for patients with congenital heart disease in the Netherlands \citep{vander2005concor}. This registry has been instrumental in advancing understanding and treatment of congenital heart conditions in the country.
Similarly, SWEDCON, the Swedish National Registry on Congenital Heart Disease, exemplifies another comprehensive approach to data collection and analysis in this specialized field \citep{bodell2017national}. By consolidating information from across Sweden, SWEDCON has become an essential tool for researchers and clinicians alike, driving improvements in patient care and outcomes.
In Belgium, the BELCODAC (BELgian COngenital Heart Disease Database combining Administrative and Clinical data) represents a more recent development \citep{ombelet2020creating}. Established in 2020, this registry takes an innovative approach by merging administrative and clinical data, providing a more holistic view of patient experiences and treatment efficacy.

While clinical registries focus on specific conditions or diseases, clinical databases cast a wider net, encompassing health information from a broader range of sources. These databases are not limited to particular conditions but instead offer a comprehensive overview of patient health data. A prime example of this broader approach is the hospital database, which typically contains a wealth of information on various medical conditions, treatments, and patient outcomes within a single healthcare institution or network.

The distinction between clinical registries and databases lies in their scope and focus. Registries are tailored to specific conditions, allowing for deep, targeted analysis and research. Databases, on the other hand, offer a more expansive view of health information, facilitating broader studies and analyses across multiple conditions and patient populations.
Both clinical registries and databases are invaluable tools in the healthcare landscape. They not only support day-to-day patient care but also drive medical research, inform policy decisions, and contribute to the overall advancement of medical knowledge and practice. As healthcare continues to evolve, these data collection and analysis systems will undoubtedly play an increasingly vital role in shaping the future of patient care and medical research. 

\section{International Classification of Disease System}

\subsection{The World Health Organisation and the International Classification of Disease System}

The International Classification of Diseases (ICD) system, maintained by the World Health Organization (WHO) \citep{hardikerterminologies}, plays a crucial role in global healthcare. This standardized coding system enables the collection of reliable data for statistical and machine learning applications, facilitating comparisons across different locations and time periods \citep{nationalcenter, raminani2015international}.
The WHO, representing 194 member states, strives to promote optimal health standards for all individuals, regardless of their socioeconomic status, race, gender, religion, or political beliefs. At its core, the organization upholds the principle that access to affordable and comprehensive healthcare is a fundamental human right.

Health is influenced by various factors, including biomedical, genetic, behavioral, socioeconomic, and environmental determinants. To measure health and wellbeing effectively, common metrics are essential. These include objective measures like life expectancy and mortality rates, as well as subjective assessments of an individual's perceived well-being.
The ICD system serves as the international standard diagnostic tool for epidemiology, health management, and clinical purposes. It translates diagnoses and health problems into alphanumeric codes, enabling systematic storage, retrieval, and analysis of data. This standardization allows for meaningful comparisons of morbidity and mortality data across hospitals, nations, and time periods.
Developed collaboratively between the WHO and international centers, the ICD system aims to group medical terms reported by physicians, medical examiners, and death certificates for statistical purposes. Its broad acceptance and official agreement for use have made it an invaluable tool for international health reporting.

The ICD system benefits various stakeholders in the healthcare sector. Clinicians, including physicians, nurses, and health workers, use it to support patient care decisions. Administrators, such as health information managers, policymakers, and insurers, rely on it for resource allocation and program management. Researchers and epidemiologists utilize the data for population health studies and the development of AI-based decision support systems.

\begin{figure*}
\centering
\includegraphics[width=\textwidth]{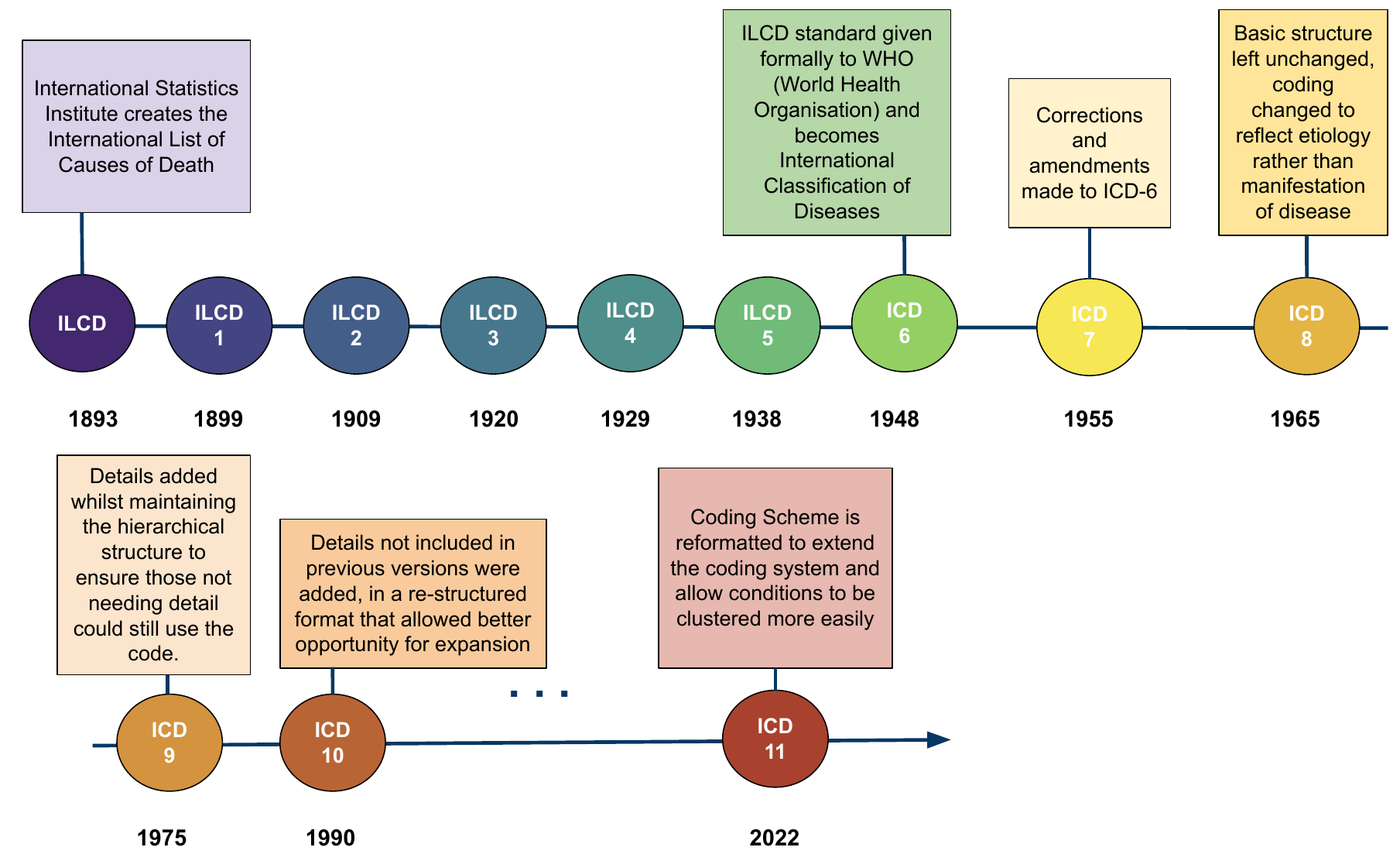}
\caption{Evolution of the International Classification of Diseases System}
\label{fig:historyICD}
\end{figure*}

\subsection{Evolution of the ICD System}
The International Classification of Diseases (ICD) system has a rich history spanning nearly four centuries, Figure \ref{fig:historyICD}. Its roots can be traced back to 17th-century England, where epidemiologist John Graunt recognized the need to organize mortality data systematically \citep{boyce2020bills}. His pioneering work, the London Bills of Mortality, marked the first statistical study of disease, though it was limited by inconsistent nomenclature and poor data quality.

The modern era of disease classification began in the late 19th century with the efforts of William Farr, often regarded as the first medical statistician. Farr's 1855 report on the nomenclature and statistical classification of diseases laid the groundwork for a more comprehensive approach, emphasizing the importance of classifying diseases based on the organic systems they affect.

In 1893, the International Statistical Institute adopted the first edition of the International List of Causes of Death, marking the birth of the modern ICD system. This classification, based on the work of Jacques Bertillon, distinguished between general diseases and those localized to specific organs or anatomical sites. The system was widely adopted by American and European countries, with an agreement to update it approximately every decade.

Throughout the 20th century, the ICD system underwent significant transformations \citep{moriyama2011history, hardikerterminologies}. The sixth revision in 1948 marked a pivotal moment as the World Health Organization (WHO) adopted it as its official classification system. This version expanded beyond mortality to include morbidity, recognizing the importance of non-fatal but chronic conditions.

Subsequent revisions brought further refinements. The ninth revision introduced a dual classification system reflecting both etiology and manifestation, while the tenth revision implemented an alphanumeric coding structure for greater flexibility and expansion.

The latest iteration, ICD-11, adopted in 2019 and effective from 2022, represents a significant leap forward \citep{lancet2019icd}. Designed for the digital era, it aims to improve diagnostic accuracy, enhance the coding of healthcare quality and safety, and better account for socioeconomic factors influencing health. ICD-11 also simplifies diagnostic descriptions, particularly in mental health.

The ICD system's evolution reflects the ongoing effort to create a comprehensive, flexible, and internationally accepted classification of diseases. Its widespread adoption has made it an indispensable tool for comparing health statistics across different countries and time periods, playing a crucial role in global health management and research.

\subsection{Understanding ICD-9 and ICD-10 Coding Systems}

\subsubsection{ICD-9 Coding System}

\begin{figure*}
\centering
\includegraphics[width=\textwidth]
{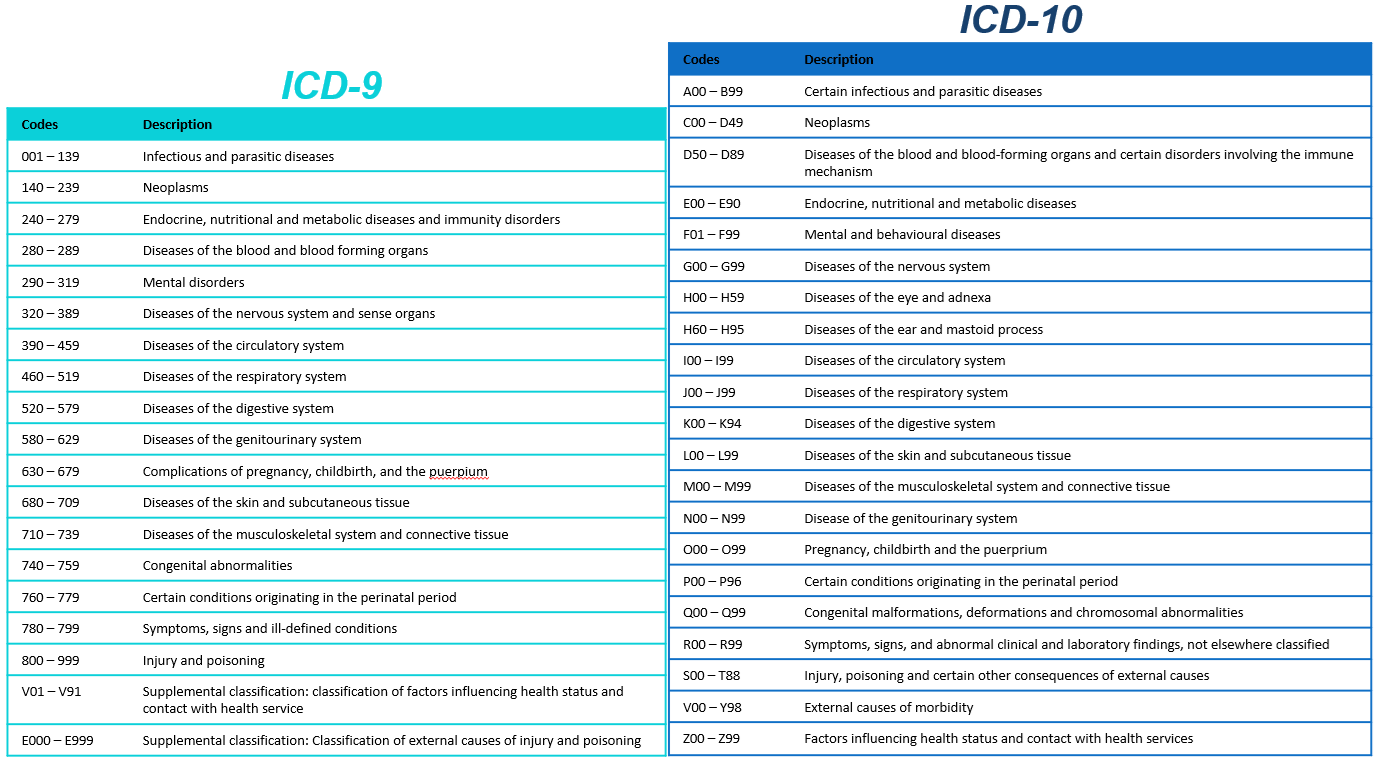}
\caption{ICD09 and ICD10 Chapters}
\label{fig:ICD9-Chapters}
\end{figure*}

The ICD-9 coding system consists of a five-digit structure: the first three digits reflect a category, while the last two digits indicate the cause or location. Categories encompass a wide range of classifications, including epidemic diseases, constitutional or general diseases, local diseases arranged by system (e.g., circulatory, respiratory), developmental diseases, and injuries. The World Health Organization mandates a minimum of three-character categories for international reporting and comparison, with the fourth digit filled with 'X' when no further subdivision information is available.

ICD-9 categories cover a broad spectrum of medical conditions, starting from epidemic diseases and progressing through various systemic disorders such as neoplasms, endocrine and metabolic diseases, blood disorders, mental disorders, and diseases of specific organ systems. The classification also includes developmental diseases, injuries, and poisoning. The last two categories use letters as the first digit and offer supplemental classifications.

\subsubsection{From ICD-9 to ICD-10}

\begin{figure}
\centering
\includegraphics[scale=0.4]
{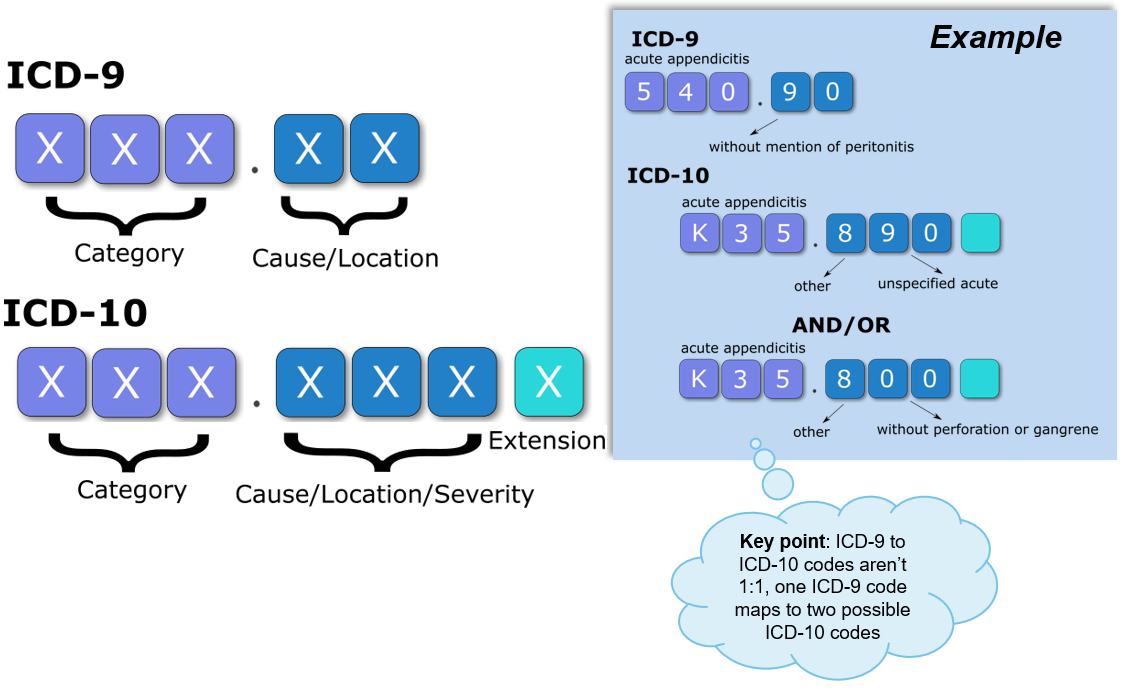}
\caption{ICD-9 VS ICD-10}
\label{fig:ICD9VSICD10}
\end{figure}

The International Classification of Diseases (ICD) has evolved significantly from its ninth revision (ICD-9) to its tenth revision (ICD-10), resulting in substantial changes in disease classification and coding, as it is shown in Figure \ref{fig:ICD9-Chapters} and Figure \ref{fig:ICD9VSICD10}. 
ICD-10 represents a major advancement over ICD-9, offering greater specificity and flexibility in disease classification \citep{world2004international}. While ICD-9 primarily used numeric codes with occasional use of letters E and V, ICD-10 adopts an alphanumeric system. The first character in ICD-10 is always alphabetic, followed by two numeric digits for the category. Subsequent characters can be either alphabetic or numeric, allowing for more detailed classification.

The structural differences between the two systems are significant as is also shown in Figure \ref{fig:ICD9-Chapters}. ICD-10 contains 21 chapters compared to ICD-9's 17, providing more comprehensive coverage of health conditions. The expanded structure of ICD-10 allows for more precise coding of diseases, injuries, and health status factors. For instance, ICD-10 separates sense organ disorders from nervous system disorders and reorganizes injury classifications by anatomical site.

One crucial aspect to understand is that there is not always a one-to-one mapping between ICD-9 and ICD-10 codes, Figure \ref{fig:ICD9VSICD10}. A single ICD-9 code might correspond to multiple ICD-10 codes, or vice versa, depending on the specificity required. This lack of direct correlation can pose challenges when transitioning between the two systems or when analyzing historical data.
ICD-10 introduces several special signs and symbols to enhance coding precision. The dagger (†) and asterisk (*) system is particularly noteworthy. The dagger code represents the underlying disease process, while the asterisk code indicates its manifestation in a particular organ or site. This dual coding allows for a more comprehensive description of complex medical conditions.
Other important symbols in ICD-10 include parentheses for supplementary information, square brackets for synonyms or alternative phrases, and colons to indicate the need for additional digits to complete a code. The terms "Not Otherwise Specified" (NOS) and "Not Elsewhere Classified" (NEC) are used to handle cases where information is incomplete or a specific code is unavailable.

The transition to ICD-10 has significant implications for healthcare information systems. It allows for more detailed patient records, improved data analysis, and better tracking of health trends. However, it also requires substantial changes in coding practices and software systems.

Looking ahead, ICD-11 has been introduced to address some limitations of ICD-10. It offers improved functionality in digital environments, incorporates more detailed morbidity information, and provides better support for multilingual applications. These ongoing developments underscore the dynamic nature of disease classification systems and their critical role in modern healthcare management.

In conclusion, understanding the differences between ICD-9 and ICD-10, as well as the special features of ICD-10, is crucial for healthcare professionals, researchers, and information systems specialists. The increased specificity and flexibility of ICD-10 offer enhanced capabilities for disease classification and health data analysis, paving the way for more precise and effective healthcare management and research. ICD-10 was mandated for use in UK healthcare system in 1995. ICD-9 is still used in publicly available EHR databases such as MIMIC III.

\section{Datasets for Research and MIMIC}

\subsection{MIMIC Critical Care Dataset: The Impact}
The MIMIC (Medical Information Mart for Intensive Care) dataset stands as a cornerstone in healthcare research, offering a comprehensive collection of de-identified electronic health records from hospital and intensive care unit settings \citep{johnson2018mimic, bulgarelli2020prediction, johnson2014mortality}. Developed by the Laboratory for Computational Physiology at MIT, MIMIC has evolved through several iterations.
This dataset's significance lies in its volume, quality, and representation of real-world hospital data spanning over a decade. It includes a wide range of clinical information such as lab tests, medical diagnoses, vital signs, and medications, providing researchers with a rich source of complex physiological signals and clinical data.

The creation of MIMIC was driven by the need to generate new knowledge from the vast amount of data captured daily in intensive care units. By making this de-identified dataset freely accessible under a data use agreement, MIMIC has opened up unprecedented opportunities for both academic and industrial research in healthcare and related fields \citep{saeed2002mimic, saeed2011multiparameter, johnson2016mimic}.
MIMIC's impact on research has been substantial and far-reaching. Since its initial release, the dataset has garnered thousands of citations, with an exponential growth in its use across various disciplines. While its primary impact has been in medicine and critical care, MIMIC has also significantly influenced research in computer science and in particular in machine learning and natural language processing.
The dataset's comprehensive nature has enabled sophisticated research in deep learning models, clinical decision support systems, and big data analytics in healthcare. It has become a valuable resource for developing and validating new algorithms and technologies in medical informatics.

Moreover, MIMIC has set a standard for de-identifying free-text and clinical information, serving as a model for other clinical databases. Its public availability has facilitated reproducibility in clinical studies, a crucial aspect of advancing medical research.

In conclusion, the MIMIC dataset represents a significant advancement in healthcare data accessibility and has become an indispensable tool for researchers across multiple disciplines. Its continued growth and impact underscore its importance in driving innovation in healthcare analytics and clinical research.

\subsection{An overview of MIMIC III: Data Linkage}


Designing a database schema to encode the complexity of EHR information presents challenges cite \citep{moody2009physionet}. The database must be accessible to doctors and other healthcare providers simultaneously and in a unified manner. Interoperability is crucial, involving enhanced quality, efficiency, and effectiveness of the healthcare system. Information should be provided in the appropriate format when needed, eliminating unnecessary duplication.

\begin{figure*}
\centering
\includegraphics[width=\textwidth]{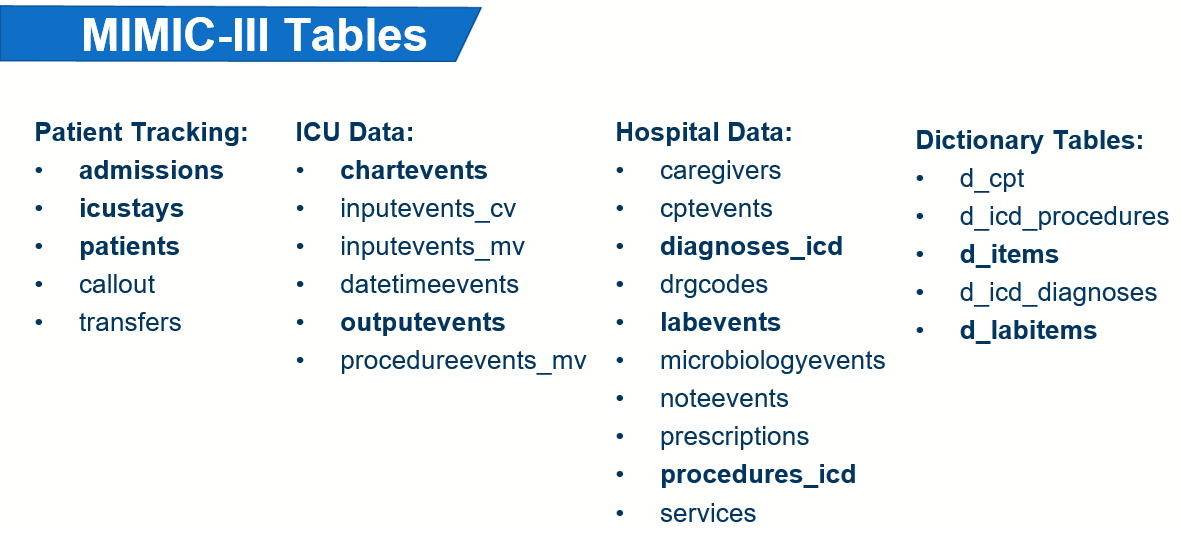}
\caption{MIMIC III as a relational database - Tables}
\label{fig:mimicIII-tables}
\end{figure*}

The choice of database and its schema architecture influences the effective management of medical data, flexibility, scalability, query performance, and interoperability. Non-proprietary standardized models are essential for building electronic health record systems that meet interoperability requirements. MIMIC-III serves as an excellent example in this regard, Figure \ref{fig:mimicIII-tables}.



During ICU stays, various signals are monitored, including vital signs, waveforms, alarms, fluids, medications, and progression reports from doctors. Hospital data includes billing details, International Classification of Disease codes related to pathology and symptoms, patient demographics, and other notes concerning medical images and discharge summaries.

A key step in building the MIMIC-III database was the de-identification process, ensuring compliance with the Health Insurance Portability and Accountability Act. All fields related to patient identification were removed, including names, telephone numbers, and addresses. Dates were shifted into the future by a random offset for each patient, preserving intervals while protecting privacy. This shift maintained seasonality, time of day, and day of the week. Special measures were taken to protect individuals with rare age characteristics, and text fields like diagnostic reports and physician notes were carefully processed to remove any protected health information.

In conclusion, MIMIC-III links data across hospitals, ICU units, and Death Registries. It provides a wealth of information, including lab examinations, medications, International Classification of Disease coding, and vital signals. As the only freely available database of its kind without major usage restrictions, MIMIC-III represents a valuable resource for healthcare research and analysis.


\subsection{MIMIC-III as a Relational Database}

This section explores the structure of MIMIC-III as a relational database, with the aim of providing an understanding of the database schema, its basic tables, Figure \ref{fig:mimicIII-tables}, and navigation methods. The MIMIC-III database contains over 53,000 distinct hospital admissions for patients aged 60 years and above, admitted to critical care between 2001 and 2012. It also includes data on more than 7,800 neonates from 2001 to 2008.

To effectively extract information from this electronic health record database, it's crucial to understand both the database schema and the data sources. Most queries will involve joining information between the basic tables holding data for patient admissions in the hospital and ICU stays.

The MIMIC-III is a relational database comprises 26 tables, reflecting the inherent hospital information sources. Its structure closely represents the raw data, with only minor adjustments made based on user feedback. These tables can be categorized into four groups: patient tracking, ICU data, hospital data, and dictionary tables. The subject ID refers to a unique patient, the hospital admission ID to a unique hospital admission, and the icustay ID to a unique ICU admission.

The core tables of MIMIC-III are the patients table, the admissions table, and the icustays table. These tables form the foundation for extracting various summary statistics. The patients table contains a subject ID identifier linking it to the admissions and ICU tables. It includes information such as date of birth (shifted for patients over 89 years old to protect privacy) and different versions of date of death. The admissions table records every unique hospitalization for each patient, including admission and discharge times, type of admission, and diagnosis information. The icustays table documents every unique ICU stay, with details like care unit, ward locations, and length of stay.


Other significant tables include the transfers table, which provides information on patient movements within the hospital, and the callout table, which records ICU discharge clearance and actual discharge times. The chartevents table is crucial for ICU data, containing all chart event observations for patients. The outputevents table focuses on patient output measurements.

When working with ICU data, particularly chartevents data, it's essential to consider the dictionary tables. These provide definitions for identifiers, allowing researchers to cross-reference codes against their respective definitions.

In the hospital data category, frequently used tables include the labevents table, containing laboratory test results. It's worth noting that in cases of disagreement between measurements, labevents should be considered the ground truth. The 'procedures\_icd' and 'diagnoses\_icd' tables contain procedure and diagnosis information coded using the International Statistical Classification of Diseases (ICD-9) system.

Researchers are advised to develop database views and transform data as needed rather than combining tables within the MIMIC data model. This approach maintains the integrity of the original data sources while allowing for flexible analysis.


\subsection{MIMIC-III - Descriptive Statistics}
 Descriptive statistics serve as a powerful tool in retrospective studies, allowing researchers to overview historic data and explain trends. This section explores how to formulate simple yet effective queries using the MIMIC-III database. It's important to note that extracting patient data can yield different estimations depending on which table identifier is used, underscoring the necessity of thoroughly understanding the database schema and its data encoding methods.

Descriptive statistics in MIMIC-III typically involve common queries that provide insights into the distribution of data within the database. These queries, while relatively simple, offer powerful ways to interrogate the database and generate intuitive summary visualizations. They often focus on patient characteristics, intensive care unit utilization, and patient outcomes such as mortality.

When estimating the number of patients, multiple approaches can lead to similar but not identical results. For instance, one can examine the number of distinct patients across care units, unique hospital admissions, or unique admissions to ICUs. Each method yields slightly different results, as some patients may have multiple admissions or transfers between units.

A fundamental query in electronic health record databases involves examining age distribution across different departments. In MIMIC-III, this is particularly useful for understanding age distribution across intensive care units. By combining the Patients and Icustays tables, researchers can compute age by subtracting the time of ICU admission from a patient's date of birth. The first\_careunit identifier in the Icustays table provides information about different care units.

Similarly, gender distribution across critical care units can be extracted by combining the Patient and Icustays tables based on the subject\_id identifier. This query collects information about the ICU each patient was admitted to, along with gender, date of birth, and hospital admission time. These data points allow for age computation and filtering of patients older than 16 years, focusing the analysis on the adult population.

Visualizing these statistics can provide insights into which conditions that affect males and females differently across various critical care units. For instance, observations might reveal higher numbers of men in intensive care units related to cardiac problems and trauma surgery.

\begin{figure*}[hbtp]
\centering
\includegraphics[width=1\textwidth]{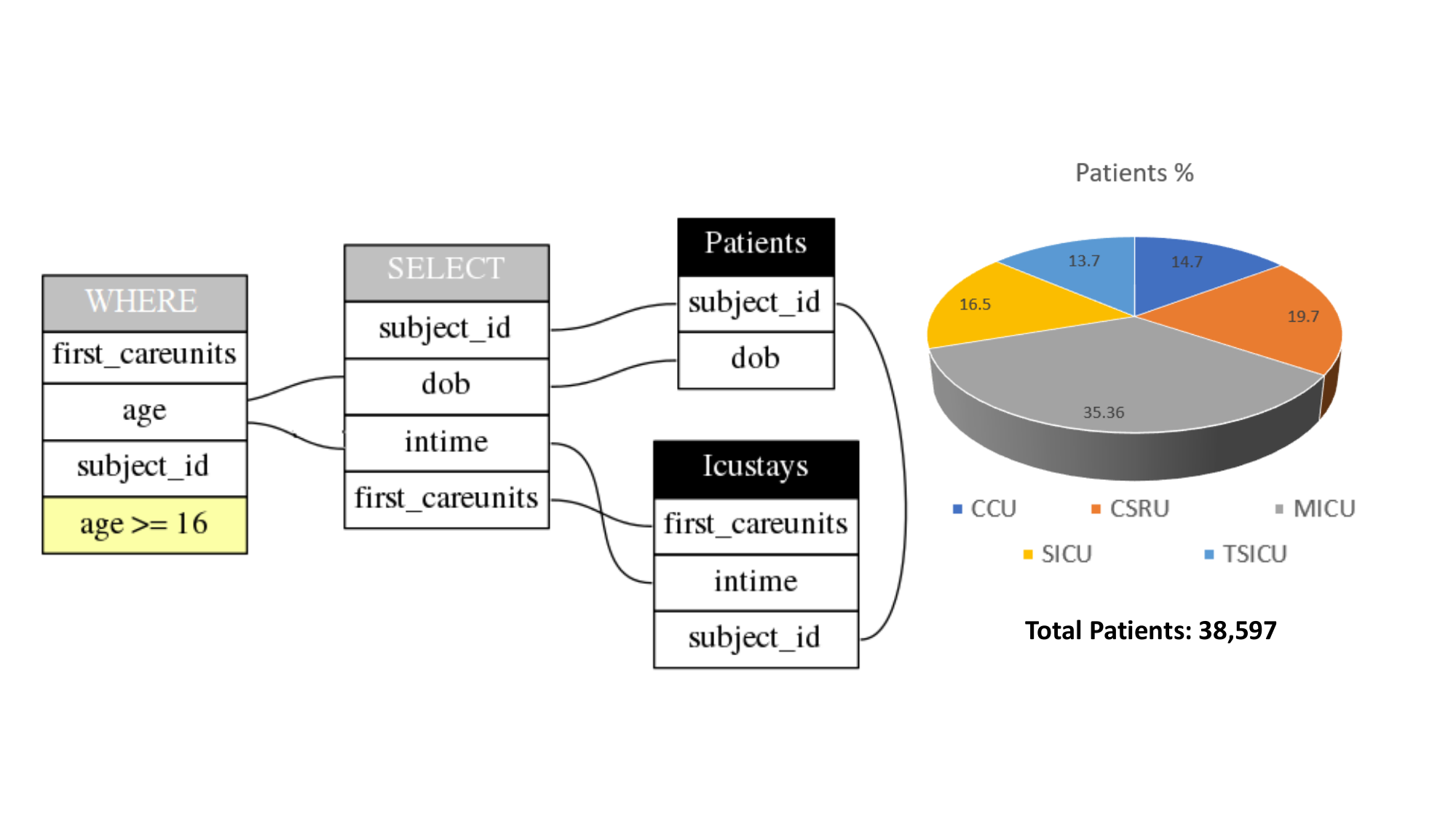}
\caption{MIMIC-III: Distinct Patients’ Hospitalisations}
\vspace{-40pt}
\label{fig:mimicIII-ExDistinctPat}
\end{figure*}

Another valuable descriptive statistic involves extracting distinct patient hospitalizations, Figure \ref{fig:mimicIII-ExDistinctPat}. This query combines the Patient and Icustays tables to obtain each patient's unique identifier and corresponding ICU admission information. It's worth noting that the Icustays identifier groups all ICU admissions within 24 hours of each other, meaning a patient transferred between ICU types might have the same Icustay ID.

\begin{figure*}[hbtp]
\centering
\includegraphics[width=1\textwidth]{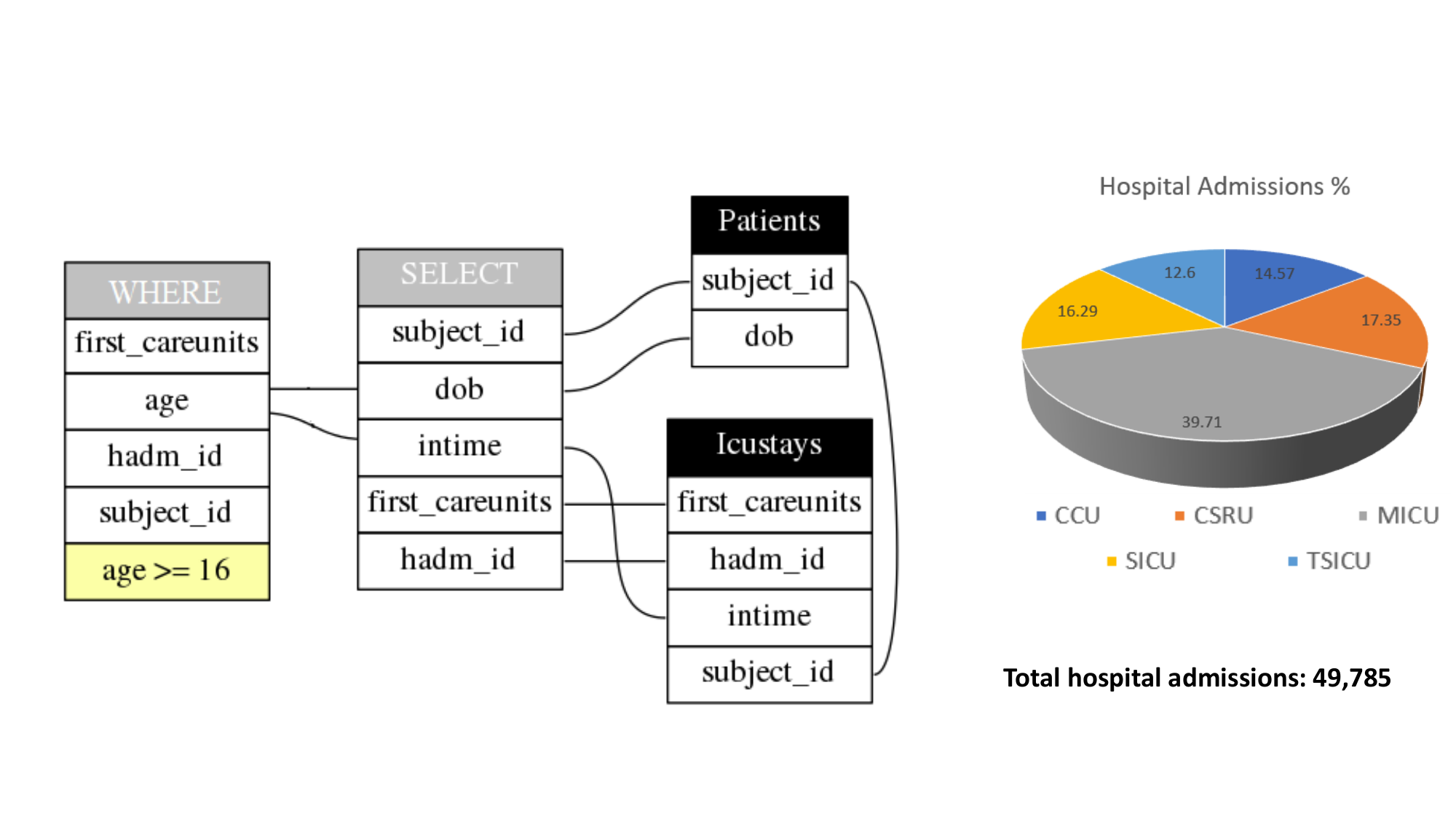}
\caption{MIMIC-III: Hospital Admissions}
\label{fig:mimicIII-ExHospitalAdmis}
\end{figure*}

Researchers might also be interested in the number of unique hospital admissions and their distribution across intensive care units, Figure \ref{fig:mimicIII-ExHospitalAdmis}. This query combines the Patients and Icustays tables based on each patient's unique subject identifier, collecting information about patient identifiers, dates of birth, and unique hospital stay identifiers.

\begin{figure*}
\centering
\includegraphics[width=1\textwidth]{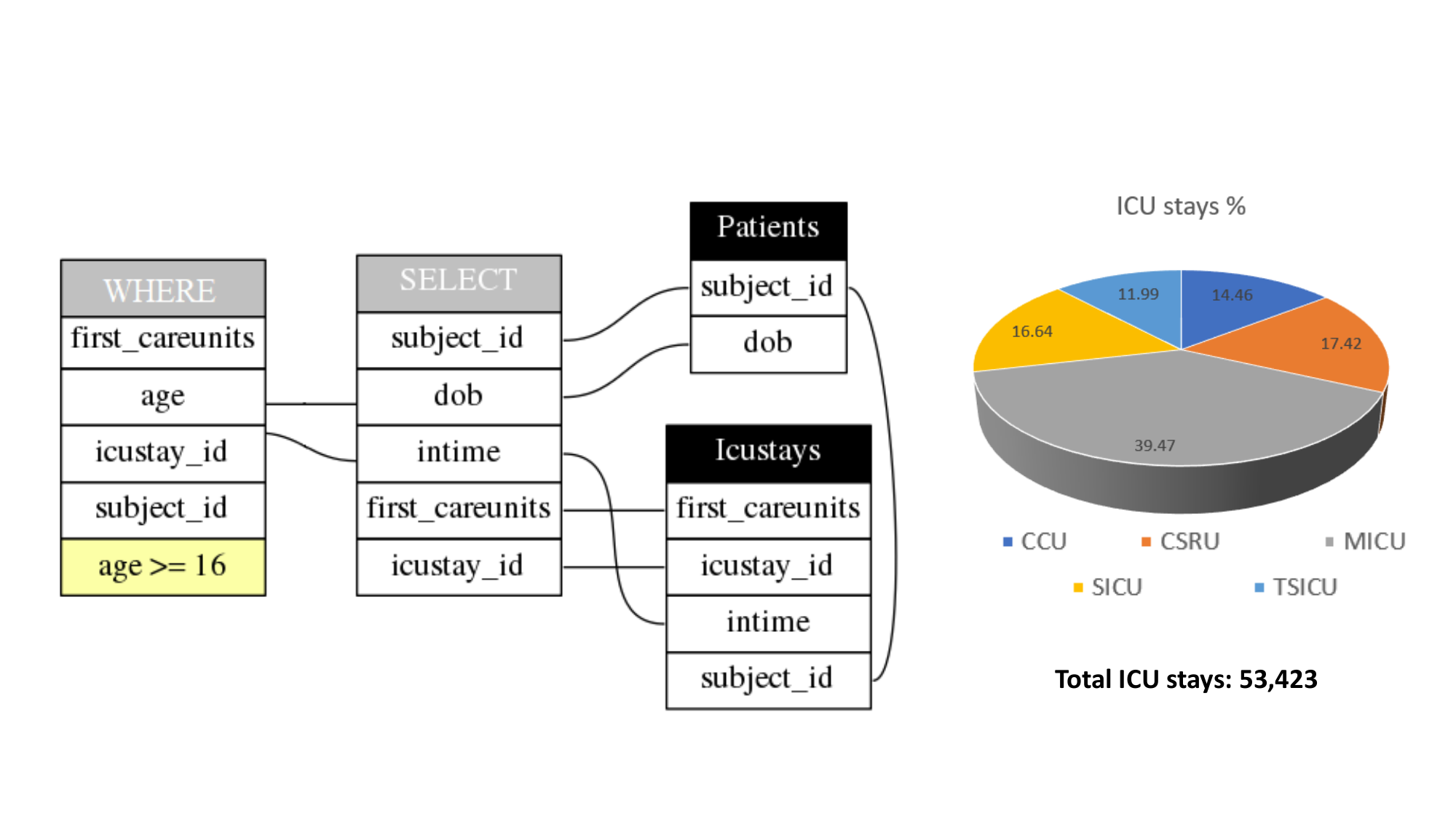}
\caption{MIMIC-III: Distinct ICU Admissions}
\label{fig:mimicIII-ExampleICUStays}
\end{figure*}

To identify distinct ICU admissions, a similar process is followed, but using the unique ICU stay identifier instead of the hospitalization identifier, Figure \ref{fig:mimicIII-ExampleICUStays}. This distinction is important because the MIMIC-III database was constructed by linking hospital and ICU information, resulting in unrelated identifiers between the two systems.

Comparing the results of these queries highlights the differences between unique patient admissions, unique hospital admissions, and unique ICU stays. Generally, the number of distinct patients across all care units is slightly less than the number of hospitalizations, which in turn is less than the number of ICU admissions across different units.

Descriptive analytics can provide valuable insights about patient numbers and resource utilization, revealing trends and anomalies over time. For example, studies based on descriptive statistics have shown steady increases in chronic illnesses like diabetes mellitus and hypertension, as well as an increasing proportion of elderly patients and women in hospital populations.

These analytics can also shed light on how emerging technologies impact patient outcomes, length of stay, associated risks of complications, and resource consumption. They can reveal trends such as declining hospital mortality from acute coronary syndromes or the increasing age of admitted patients over time.

While descriptive analytics offer substantial information about historic data and can explain trends, they are limited to retrospective studies. They cannot predict future outcomes or prevent diseases and high mortality rates. Nevertheless, they remain a crucial tool in understanding the complex landscape of critical care and hospital utilization patterns.

\subsection{MIMIC III and Clinical Outcomes: Mortality and Length of Stay}

\subsubsection{Mortality as a clinical outcome}

Mortality is the primary outcome measure employed in prediction studies across critical care and much of medicine. As a reliable indicator (surrogate) of illness severity, clinicians and researchers use this signal to identify trends and correlations between clinical data and patient outcomes. However, defining mortality is more complex than it might initially appear, particularly when dealing with retrospective data.

In the MIMIC database, complications arise from situations such as patients whose consecutive hospitalisations ultimately result in mortality. While these are considered as distinct hospital admissions, the underlying cause is the same. Mortality data is also prone to selection bias depending on the information source. For instance, hospital mortality records might not include deaths occurring after discharge to home care, potentially biasing any trained models if there's a routine process for discharging patients with severe illnesses to home care.

Researchers have some control over how mortality is defined as an outcome. The definition of mortality as death during the first 30 days of admission provides a more robust signal for immediate physiological abnormalities associated with the patient's admission. Conversely, one-year mortality emphasizes chronic illnesses and parallel conditions.

\begin{figure*}
\centering
\includegraphics[width=\textwidth]{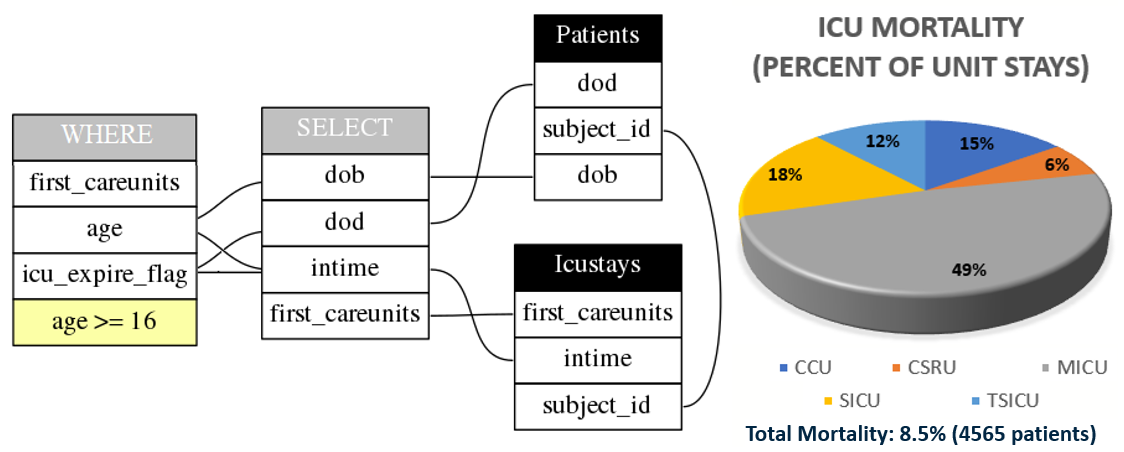}
\caption{MIMIC-III: ICU Mortality as a clinical outcome}
\label{fig:mimicIII-OutcomeICUMortality}
\end{figure*}

To extract mortality numbers for adult ICU patients in MIMIC, Figure \ref{fig:mimicIII-OutcomeICUMortality}, we combine the Patients and 'Icustays' tables using the 'subject\_id'. We extract the date of death, date of birth, ICU transfer time, and the type of ICU first admitted to. A patient is considered to have died in the ICU if their death was registered while in the ICU, or within six hours before admission or after leaving. This query focuses on patients older than 60 years.
Visualizing these results shows that total ICU mortality is about 8.5\%, corresponding to approximately 4,565 patients. The medical ICU accounts for 49\% of these deaths, despite having only about 39\% of total ICU patients, indicating a higher mortality rate in this unit compared to others.

\begin{figure*}
\centering
\includegraphics[width=\textwidth]{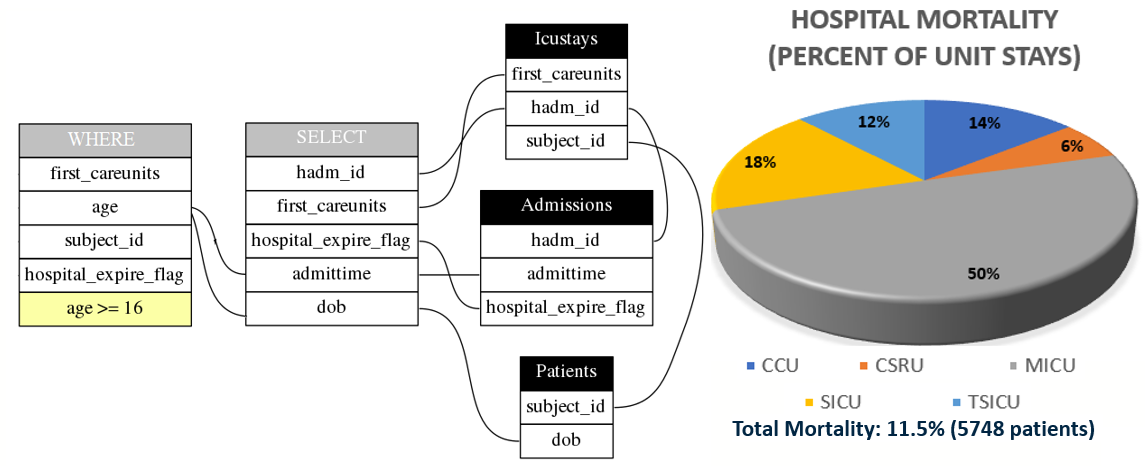}
\caption{MIMIC-III: Hospital Mortality as a clinical outcome}
\label{fig:mimicIII-OutcomeHospitalMortality}
\end{figure*}

For hospital mortality, we use the hospital\_expire\_flag in the admissions table, Figure \ref{fig:mimicIII-OutcomeHospitalMortality}. This query combines the ICU stays, patients, and admissions tables, filtering for patients older than 16. The results show an increase in total mortality to 11.5\% when considering hospital-wide deaths, with 50\% occurring in the medical ICU.

\begin{figure*}
\centering
\includegraphics[width=\textwidth]{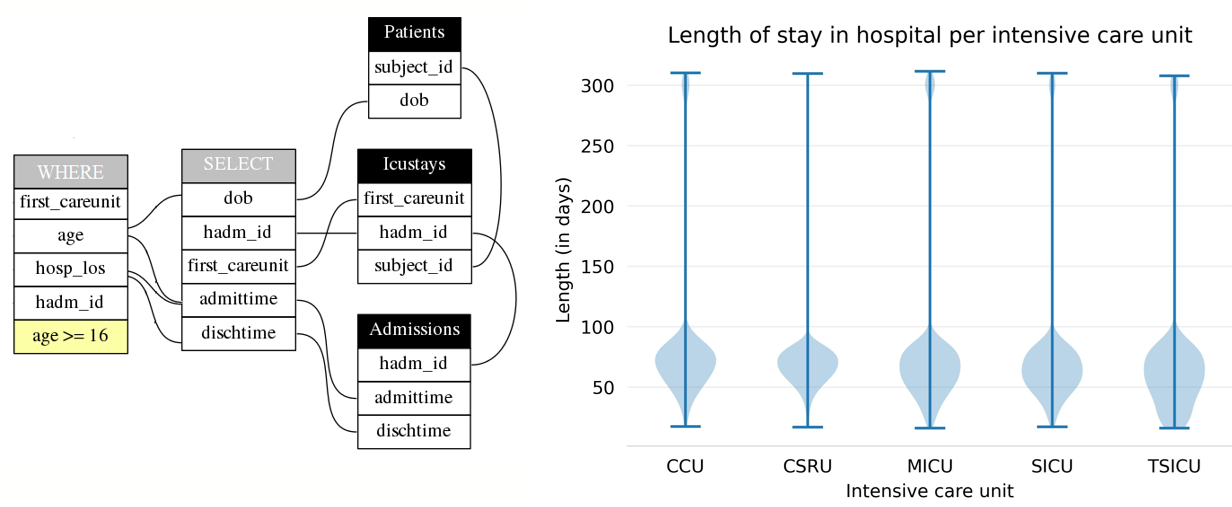}
\caption{MIMIC-III: Hospitalisation length as a clinical outcome}
\label{fig:MIMIC-III-HospitalisationLength}
\end{figure*}

\subsubsection{Length of Stay as a clinical outcome}
Length of stay is another crucial metric, relating to both patient outcomes and resource utilization. To extract this data, we combine the patients, ICU stays, and admissions tables, focusing on patients older than 60. Length of stay is computed by subtracting the discharge time from the admission time, using the unique hospital admission identifier to ensure each admission corresponds to only one ICU stay, Figure \ref{fig:MIMIC-III-HospitalisationLength}.

In conclusion, while mortality and length of stay are the most common outcomes used in predictive models, their seemingly simple definitions can be deceptive. Small differences in estimation methods can lead to significantly different conclusions. The MIMIC-III database provides a valuable resource for understanding how to form queries and extract this critical data, offering researchers a robust platform for developing and refining predictive models in critical care settings.

\subsection{MIMIC III: Exploring Patient Data - Vital Signs Extraction for a Single Patient}

Understanding vital signs and medication data for individual patients admitted to intensive care units is crucial, as it forms the foundation for extracting clinical variables across multiple patients. These variables can include lab exams, signs and waveforms, as well as doctor reports and prescriptions.

\begin{figure*}
\centering
\includegraphics[width=\textwidth]{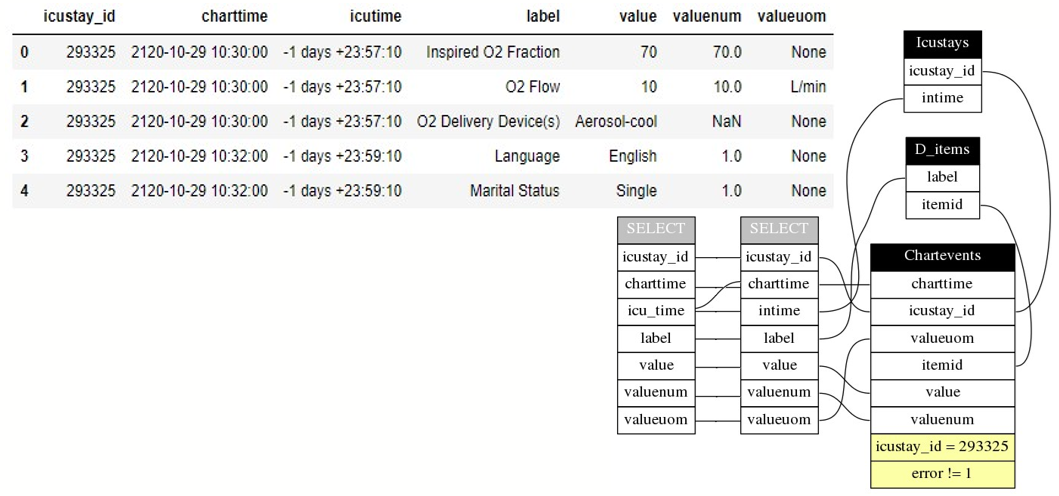}
\caption{MIMIC-III: Charted Events for a Single Patient}
\label{fig:mimicIII-chartedEventsSP}
\end{figure*}

To begin exploring patient data, we start by selecting a random ICU stay identifier and its corresponding subject identifier. This allows us to focus on data related to a specific patient and ICU admission. By combining the Admissions, ICU Stays, and Patients tables, we can create a comprehensive view of the patient's stay, including their unique subject identifier, hospital identifier, admission type, diagnosis, ICU stay identifier, first and last care units, and ICU admission time, Figure \ref{fig:mimicIII-chartedEventsSP}.

The Chartevents table is a key source of data obtained from patients in the ICU. It contains information from both CareVue and MetaVision ICU databases, displaying routine vital signs and additional care-relevant information such as ventilator settings, laboratory values, code status, and mental status. To extract all charted events for a single patient, we combine the Chartevents table with ICU stays and the Dictionary Items table. This allows us to retrieve the time of charted events, the time elapsed since ICU admission, event labels, values, and measurement units.

\begin{figure*}
\centering
\includegraphics[width=\textwidth]{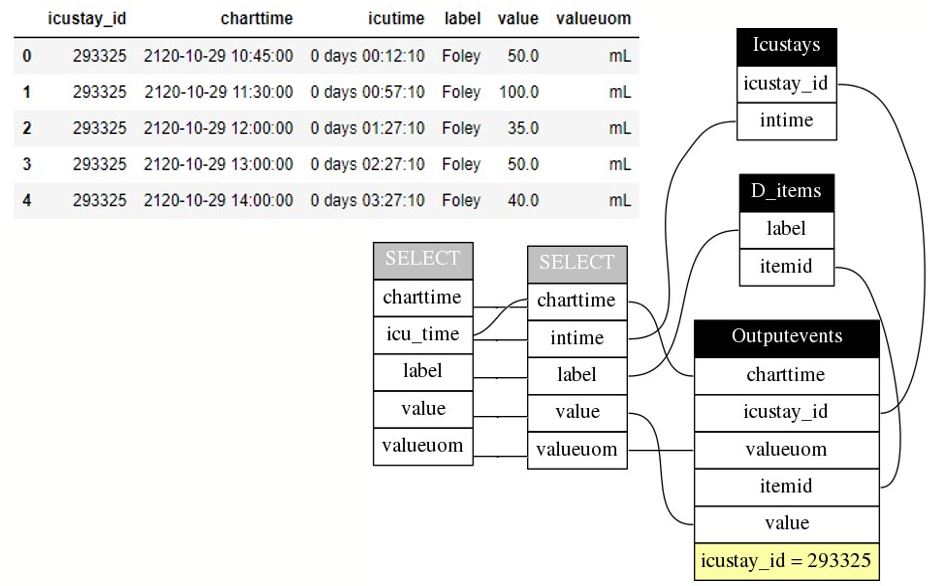}
\caption{MIMIC-III: Output Events of ICU Stay for a Single Patient}
\label{fig:mimicIII-OutputEventsSP}
\end{figure*}

Similarly, the Outputevents table provides crucial data on body fluids and other measurements, such as those from Foley catheters. We use a similar process to extract this data, joining the ICU stays table with the Outputevents table and the Dictionary table for item verification, Figure \ref{fig:mimicIII-OutputEventsSP}.

\begin{figure*}
\centering
\includegraphics[width=0.5\textwidth]{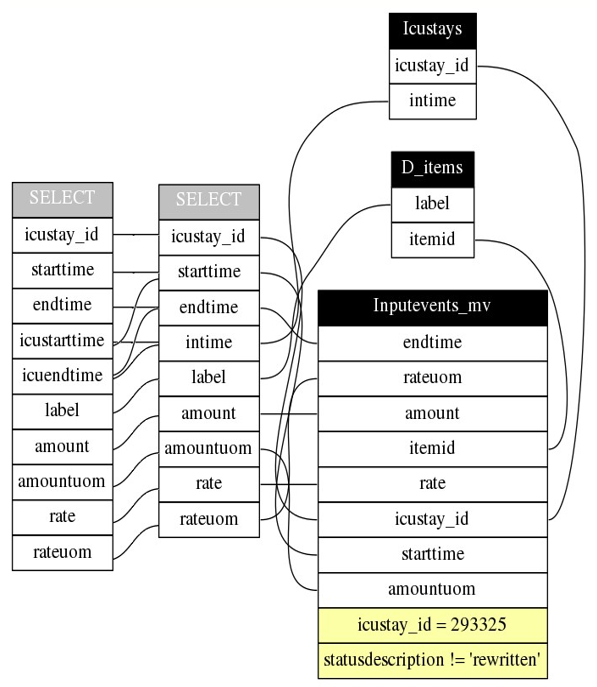}
\caption{MIMIC-III: Medication intakes during an ICU stay for a single patient}
\label{fig:mimicIII-MedicationSP}
\end{figure*}

Medication intake during an ICU stay is recorded in the Inputevents\_mv table. By joining this table with ICU stays and Dictionary Items, we can collect all input events for a single patient's ICU stay, including start and end times, labels, amounts, measurement units, and administration frequency, Figure \ref{fig:mimicIII-MedicationSP}.

\begin{figure*}
\centering
\includegraphics[width=0.5\textwidth]{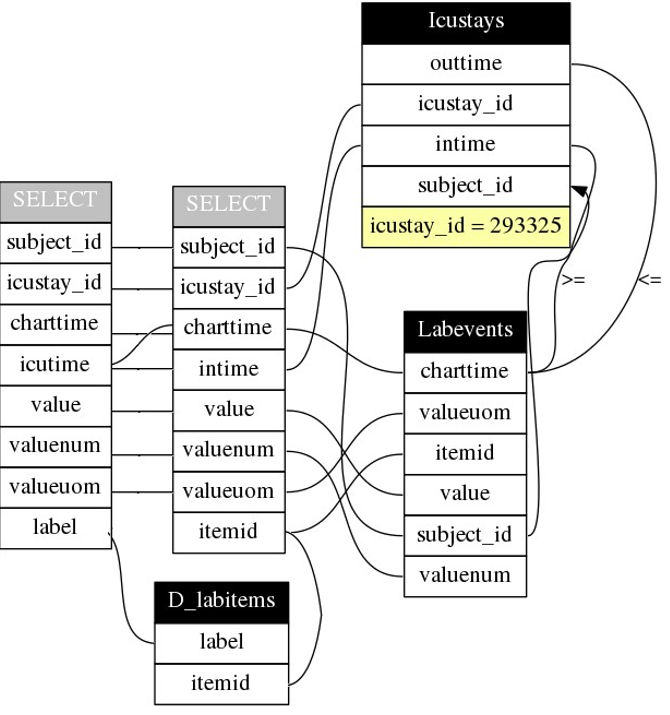}
\caption{MIMIC-III: Lab events during an ICU stay for a single patient}
\label{fig:mimicIII-LabEventsSP}
\end{figure*}

Lab events, which are part of the hospital database rather than the ICU stays database, are summarized in the Labevents table, Figure \ref{fig:mimicIII-LabEventsSP}. These events might include measurements such as red blood cell count or magnesium levels. It is important to note that when there is a discrepancy between Labevents and Chartevents, the Labevents table should be considered the ground truth.

\begin{figure*}
\centering
\includegraphics[width=0.8\textwidth]{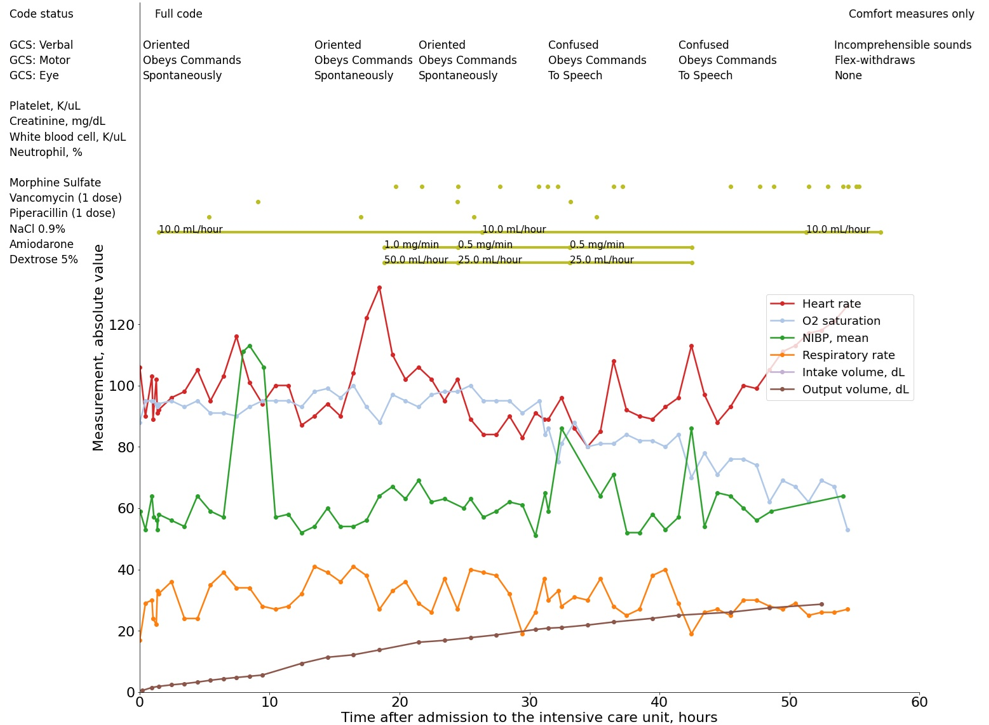}
\caption{MIMIC-III: Vital signs during an ICU stay for a single patient}
\label{fig:mimicIII-VitalSignsSP}
\end{figure*}

Visualizing this extracted data provides a comprehensive view of the patient's condition over time. We can plot various measurements against the time since ICU admission, typically in hours. This visualization might include the Glasgow Coma Scale (which measures consciousness level), lab event results, medication administration (both frequent and sporadic), and vital signs such as heart rate, O2 saturation, respiratory rate, and output volume. An example is shown at Figure \ref{fig:mimicIII-VitalSignsSP}.

These visualizations allow clinicians to understand a patient's progress in the intensive care unit over time. They provide a holistic view of the patient's condition, combining various data points to create a comprehensive picture of the patient's health status and treatment.

In summary, extracting data from MIMIC-III for a single patient in a single ICU stay involves examining charted events (primarily vital signs), output events (secretions and urine), input events (administered medications), and lab events (laboratory tests). This detailed exploration of individual patient data sets the stage for broader analyses across multiple patients, enabling researchers and clinicians to identify patterns, trends, and potential areas for improving patient care in intensive care settings.

\newpage
\subsection{MIMIC III and ICD Codes: ICD-9 in MIMIC-III}

The International Classification of Diseases, 9th Revision (ICD-9) serves as the disease classification system used in the MIMIC-III database. This section demonstrates how ICD-9 codes can be utilized to extract summary statistics from the MIMIC-III database, such as the number and distribution of patients across age groups diagnosed with specific diseases.

\begin{figure*}[h]
\centering
\includegraphics[width=\textwidth]{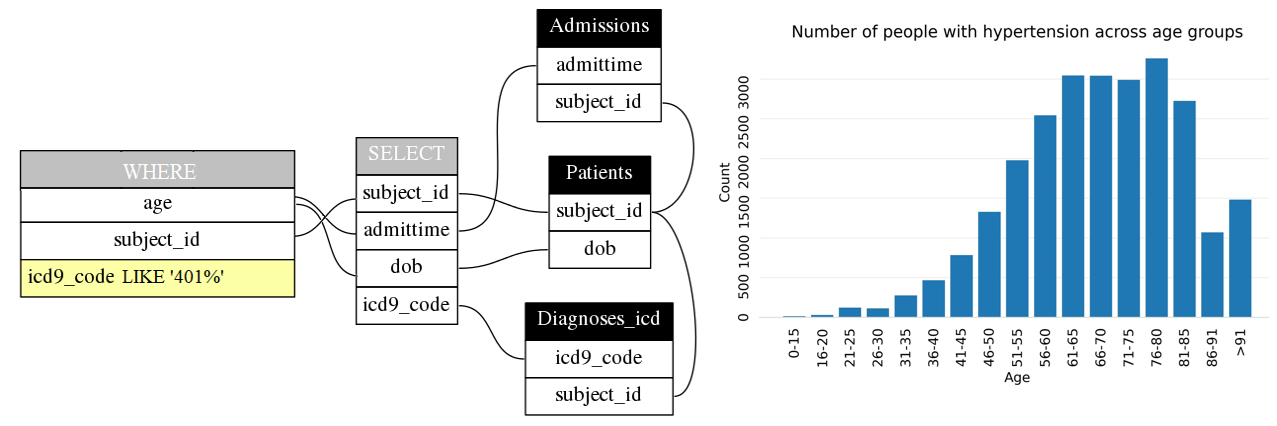}
\caption{MIMIC-III: ICD-9 Code Queries in MIMIC-III}
\label{fig:mimicIII-ICD9-hypertension}
\end{figure*}

In MIMIC-III, ICD codes are stored in the Diagnoses\_icd table, which can be linked to the Patients and Admissions tables via the subject\_id. To extract meaningful data, such as patients diagnosed with hypertension who are at least 30 years old, we need to join these tables, Figure \ref{fig:mimicIII-ICD9-hypertension}. This allows us to find the age of patients based on their date of birth and admission date, filter for age, and then identify patients assigned specific ICD-9 codes related to hypertension.
Visualizing these results can provide valuable insights. For instance, a histogram of age versus the number of patients with hypertension typically shows a peak between 60 and 80 years old, aligning with clinical expectations.

\begin{figure*}
\centering
\includegraphics[width=0.5\textwidth]{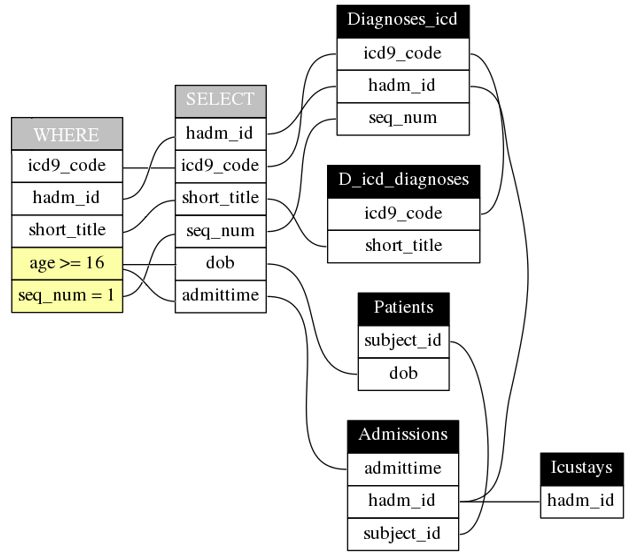}
\caption{MIMIC-III: Extracting the distribution of ICD-9 Codes across ICUs}
\label{fig:mimicIII-ICD9-QueryAcrossICUs}
\end{figure*}

Another useful query involves extracting the distribution of ICD-9 codes across intensive care units in the MIMIC-III database. This process involves sorting ICD-9 codes by prevalence, selecting the top five codes, and focusing on patients older than 60 years admitted to an ICU. By combining the patient, admission, ICU stays, and diagnosis tables, we can extract information such as patient age, primary diagnosis, ICD-9 code description, and hospital admission identifier, Figure \ref{fig:mimicIII-ICD9-QueryAcrossICUs}.

\begin{figure*}
\centering
\includegraphics[width=1.0\textwidth]{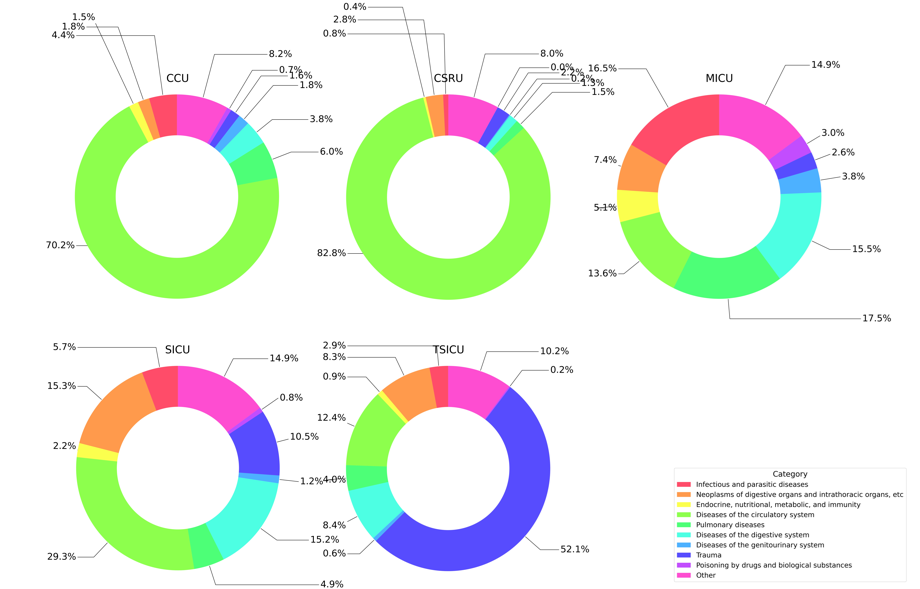}
\caption{MIMIC-III: The distribution of ICD-9 Codes across ICUs}
\label{fig:mimicIII-ICD9-AcrossICUs}
\end{figure*}

Visualizing these results across different ICU types reveals interesting patterns, Figure \ref{fig:mimicIII-ICD9-AcrossICUs}. For example, in coronary care units and cardiac surgery recovery units, diseases related to the circulatory system are most prevalent. Medical ICUs show a mix of circulatory and pulmonary diseases, while trauma surgery ICUs, as expected, feature more trauma-related diagnoses. Surgical and medical ICUs tend to display a more diverse profile of ICD codes.

It's important to note that a single hospital admission can be assigned multiple ICD-9 codes, reflecting the fact that a patient may suffer from several diseases simultaneously upon admission. When querying, attention should be paid to the priority of each ICD-9 code to focus on primary diagnoses.

In summary, understanding the structure of ICD-9 codes and their implementation in MIMIC-III allows researchers to extract valuable information about disease distribution across age groups and ICU types. This knowledge forms a crucial foundation for more advanced analyses and can provide important insights into patient populations and care patterns in intensive care settings.

\newpage
\subsection{Clinical Concepts in MIMIC III}

Clinical concepts in MIMIC III represent definitions and models that provide information about the severity of illness, based on pre-specified sets of clinical variables provided by field experts. These concepts were developed well before the success of machine learning models to quantify the severity of illness or organ failure, predict specific patient events, and allow for timely intervention and treatment to improve outcomes and provide better care administration.

The development of clinical concepts based on electronic health record databases is challenging and resource-intensive, requiring close collaboration between data scientists and domain experts. MIMIC provides an ideal substrate for building clinical concepts collaboratively, bridging the gap between those familiar with the database and clinical workflows, and data scientists or statisticians. These models are precursors to machine learning models, with some exploiting techniques such as regression to refine initial expert intuition.

Concepts in MIMIC include severity of illness scores, organ dysfunction scores, timing of treatment, definition of sepsis, and comorbidities. The inclusion of these concepts in the MIMIC repository helps remove barriers for data scientists unfamiliar with the clinical environment.

Severity of illness scores, such as the Acute Physiology Score (APS) and its variations, have been developed to assess patient acuity, particularly at the time of ICU admission. These scores examine disturbances in normal bodily functioning and typically comprise at least 10 independent components calculated using data from the first 24 hours of a patient's stay. The APS, also known as APACHE (Acute Physiology, Age, and Chronic Health Evaluation), has evolved through several versions, incorporating more sophisticated statistical techniques and expanding its applications to include risk stratification, mortality prediction, and benchmarking hospital performance.

The Simplified Acute Physiology Score (SAPS) aimed to streamline the APS by reducing the number of required physiological parameters. Its subsequent versions incorporated data-driven feature selection and expanded the geographical scope of the data used in its development.
Organ dysfunction scores, another important concept in MIMIC, assess multiple organ systems or focus on specific organs. Examples include the Sequential Organ Failure Assessment (SOFA), Logistic Organ Dysfunction System (LODS), Model for End-Stage Liver Disease (MELD), and various kidney disease assessment tools. These scores are used for prediction and prognosis, following a similar development pathway to illness severity scores.

The timing of treatment and sepsis detection are crucial concepts implemented in MIMIC. Sepsis, a life-threatening organ dysfunction caused by a dysregulated host response to infection, is a major concern in ICUs. MIMIC III provides case selection for sepsis based on ICD-9 codes, enabling the development of early warning systems using high-resolution physiological data and machine learning algorithms.

Comorbidities, referring to chronic conditions that patients have prior to ICU admission, are another important concept in MIMIC. The database defines 29 categories based on ICD-9 codes, recognizing the impact of these conditions on a patient's probability of surviving significant illness.

While these clinical concepts provide valuable tools for population studies and quality of care assessments, their predictive power for individual patient outcomes is limited. This limitation stems partly from variant disease definitions and patient population heterogeneity. Advanced machine learning methods promise to deliver better warning systems and potentially outperform these traditional clinical concepts.
In summary, concepts in MIMIC III implement well-known scores of illness severity and organ dysfunction, providing an ideal substrate for comparing their performance and enabling reproducibility. However, these concepts are primarily based on expert knowledge, and their prognostic power is limited. With the availability of big data, there is hope that state-of-the-art machine learning approaches will provide personalized outcome predictions, potentially outperforming the clinical concepts developed over the past few decades. This evolution in clinical prediction models represents an exciting frontier in critical care research and practice.

\subsection{Flowchart of Patient Inclusion: Relation of Catheterization to Mortality - An Example Study}

This section discusses a clinical study that utilizes the MIMIC database to estimate the relationship between mortality and an intervention in intensive care unit patients. The study serves as an example of a complex selection pipeline for patients and clinical variables, followed by the construction of a model that enables statistical comparisons between control and intervention groups.

Indwelling arterial catheters are widely used in intensive care units for continuous hemodynamic monitoring and blood gas analysis. While these catheters provide crucial information about oxygen delivery to tissues and organs, they also carry risks such as bloodstream infections and vascular complications. A study published in CHEST in 2015 aimed to test the hypothesis of whether catheterization was associated with mortality in the ICU, prompted by concerns about the high rate of bloodstream infections associated with central venous catheters and, to a lesser extent, indwelling arterial catheters \citep{hsu2015association}.

The study's purpose was to examine the association between indwelling arterial catheters and outcomes in a large cohort of hemodynamically stable intensive care patients with respiratory failure undergoing mechanical ventilation. This cohort was extracted from the MIMIC database using specific selection criteria based on clinical expertise.

\begin{figure*}
\centering
\includegraphics[width=0.4\textwidth]{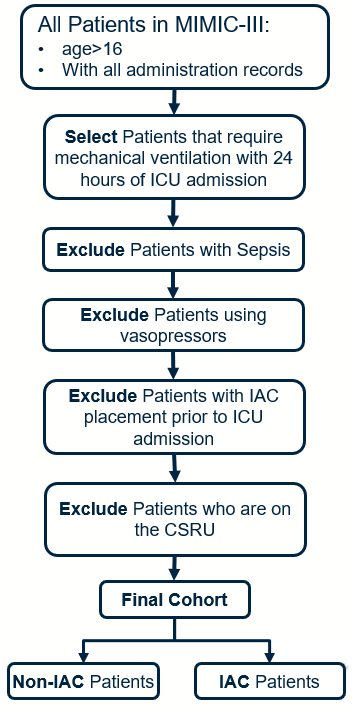}
\caption{MIMIC-III: Flowchart of Patient Inclusion and Exclusion Criteria}
\label{fig:mimicIII-flowChart}
\end{figure*}

The patient inclusion flowchart began with all patients in MIMIC III older than 16 years with administration records, Figure \ref{fig:mimicIII-flowChart}. The study focused on patients requiring mechanical ventilation within 24 hours of ICU admission. To eliminate confounding factors, several exclusion criteria were applied. Patients with sepsis were excluded due to the life-threatening nature of this condition. Those administered vasopressor drugs were also excluded, as these drugs can cause further complications by constricting blood vessels. Patients with indwelling arterial catheterization placement prior to ICU admission and those in the cardiac surgery recovery unit were also removed from the study cohort.
The final cohort was split into two groups: patients who received catheterization and those who did not, with the latter serving as the control group for comparison. This division allowed for hypothesis testing on the significance of the relationship between mortality and catheterization.

The primary outcome of the study was 28-day mortality, identifying which patients died within 28 days of ICU admission. Secondary outcomes included length of stay in the ICU and hospital, duration of mechanical ventilation, and the average number of arterial and venous blood gas measurements performed per day.

The study defined the presence of an indwelling arterial catheter as placement of the invasive catheter at any point after the initiation of mechanical ventilation. The developed model included variables reflecting patient demographics, comorbidities based on ICD-9 codes, vital signs, and laboratory results. The measurements used were those immediately preceding the onset of mechanical ventilation.

The model estimated a propensity score, representing the likelihood of catheter placement. A genetic algorithm was employed to select the most relevant clinical variables and optimize the performance of the propensity score model. The propensity score matching process paired each patient from the catheterization group with patients from the non-catheterization group, allowing for an estimation of the intervention's impact.

The study's findings revealed no significant differences in 28-day mortality between the two groups. However, patients with indwelling arterial catheterization had a lower likelihood of discharge from the ICU within 28 days.

In summary, this example demonstrates how the MIMIC database can be used to conduct complex clinical studies. It highlights the importance of careful patient selection criteria to focus on relevant patients while eliminating confounding parameters. The study also illustrates the definition of primary and secondary outcomes and the development of a propensity model to examine differences between intervention and control groups. This approach provides a framework for using large-scale electronic health record databases to investigate clinical interventions and their outcomes in critical care settings.

\begin{figure*}
\centering
\includegraphics[width=\textwidth]{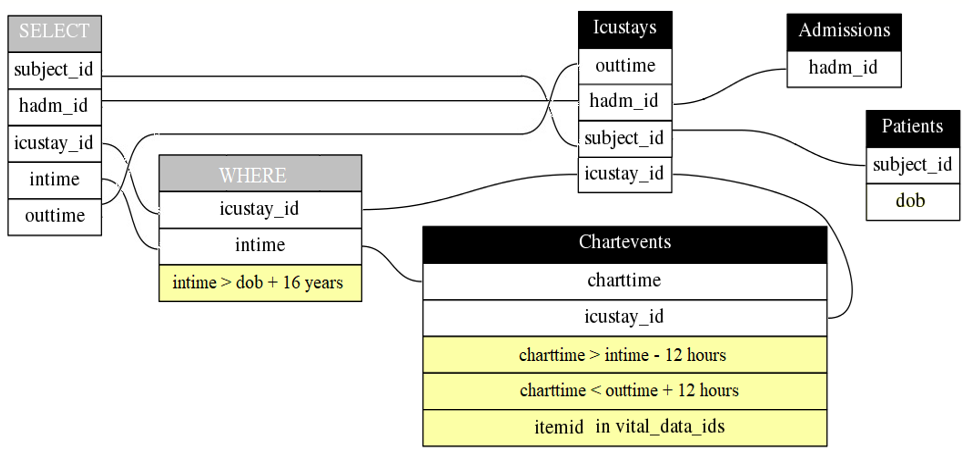}
\caption{MIMIC-III: Query to Extract All Patients at ICU and all administration records}
\label{fig:mimicIII-flowChartQuery}
\end{figure*}

\newpage
\subsubsection{Data Extraction from MIMIC-III for Catheterization and Mortality Study}

This section provides a detailed guide to extracting and processing data from the MIMIC-III database for a study investigating the relationship between catheterization and mortality in intensive care units (ICUs). We will walk through each step of the data extraction pipeline, providing specific SQL queries and explanations to ensure reproducibility.

We begin by extracting vital data for patients over 16 years old. This involves joining multiple tables in the MIMIC-III database, Figure \ref{fig:mimicIII-flowChartQuery}.
This query creates a table adult\_patients containing the ICU stay identifier, subject identifier, hospital admission identifier, ICU admission and discharge times, date of birth, and chart time for heart rate measurements.

\begin{figure*}[h]
\centering
\includegraphics[width=0.7\textwidth]{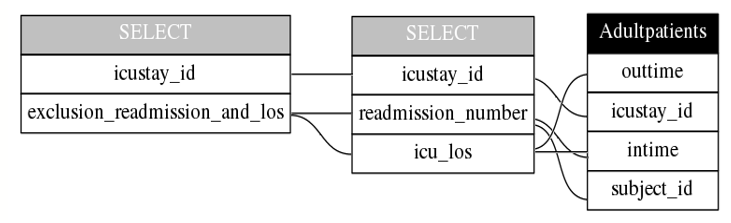}
\caption{MIMIC-III: Query to exclude re-admissions}
\label{fig:mimicIII-exludeReadmissionsQuery}
\end{figure*}

Next, we exclude readmissions and ICU stays shorter than one day, Figure \ref{fig:mimicIII-exludeReadmissionsQuery}.
This query creates a table eligible\_stays that includes only ICU stays lasting at least 24 hours and marks the admission order for each patient.

\begin{figure*}[h]
\centering
\includegraphics[width=\textwidth]{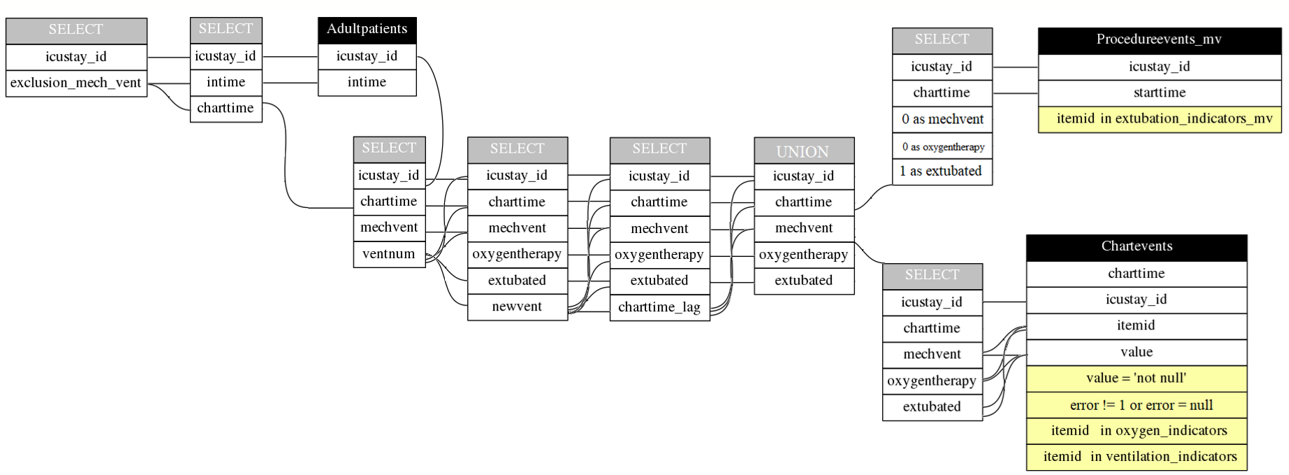}
\caption{MIMIC-III: Query to include all patients with ventilation}
\label{fig:mimicIII-patVentilationQuery}
\end{figure*}

We now identify patients who required mechanical ventilation within the first 24 hours of ICU admission, Figure \ref{fig:mimicIII-patVentilationQuery}. This query identifies patients who received mechanical ventilation within 24 hours of ICU admission.

\begin{figure*}[h]
\centering
\includegraphics[width=0.8\textwidth]{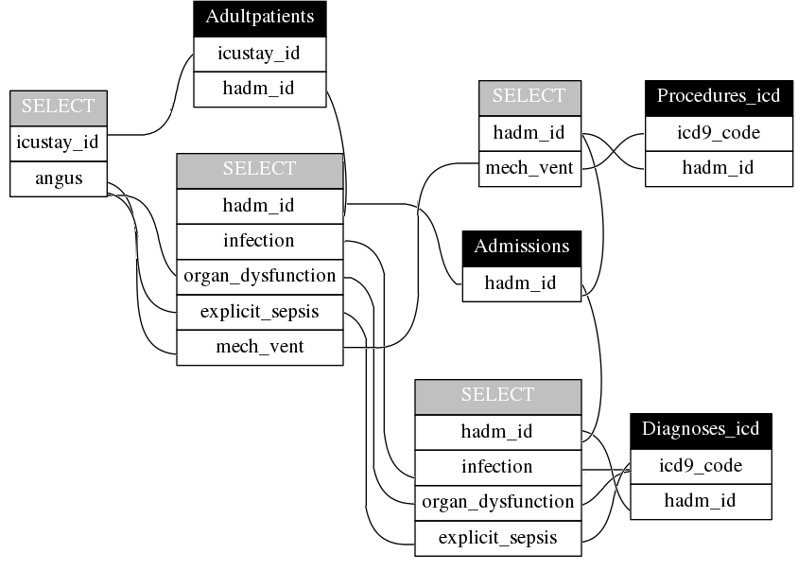}
\caption{MIMIC-III: Exclude patients with Sepsis}
\label{fig:mimicIII-excludePatSepsis}
\end{figure*}

\begin{figure*}[h]
\centering
\includegraphics[width=0.7\textwidth]{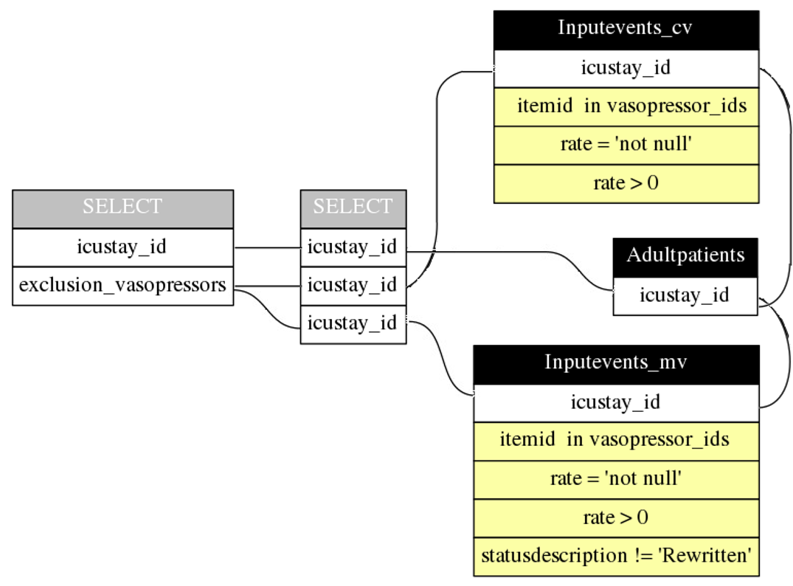}
\caption{MIMIC-III: Exclude patients with Vasopressors}
\label{fig:mimicIII-excludeVasopressors}
\end{figure*}

\begin{figure*}[h]
\centering
\includegraphics[width=0.8\textwidth]{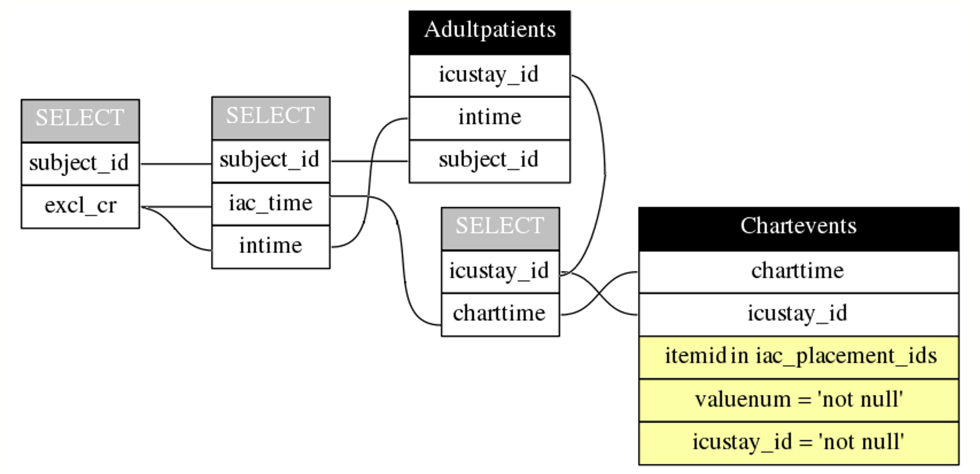}
\caption{MIMIC-III: Extract patients with IAC placed}
\label{fig:mimicIII-PatIACplaced}
\end{figure*}

\begin{figure*}[h]
\centering
\includegraphics[width=0.8\textwidth]{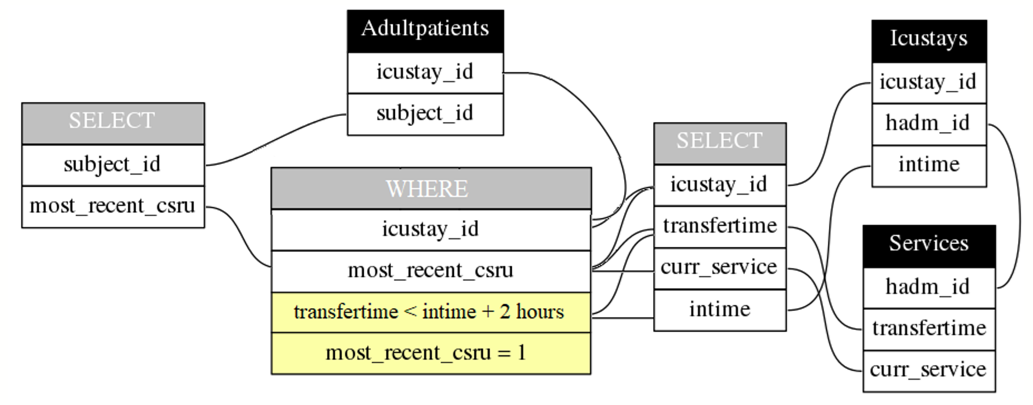}
\caption{MIMIC-III: Exclude patients in CSRU or CCU}
\label{fig:mimicIII-excludeCSRU_CCU}
\end{figure*}

We apply the following exclusion criteria:

\begin{itemize}
\item We exclude patients diagnosed with sepsis using ICD-9 codes, Figure \ref{fig:mimicIII-excludePatSepsis}.
\item we exclude patients with specific medications, Figure \ref{fig:mimicIII-excludeVasopressors}.
\item We exclude patients with catheters placed before ICU admission, Figure \ref{fig:mimicIII-PatIACplaced}.
\item Lastly, we exclude patients with prior surgery or Coronary Care Unit stays, Figure \ref{fig:mimicIII-excludeCSRU_CCU}.
\end{itemize}

\begin{figure*}[h]
\centering
\includegraphics[width=0.8\textwidth]{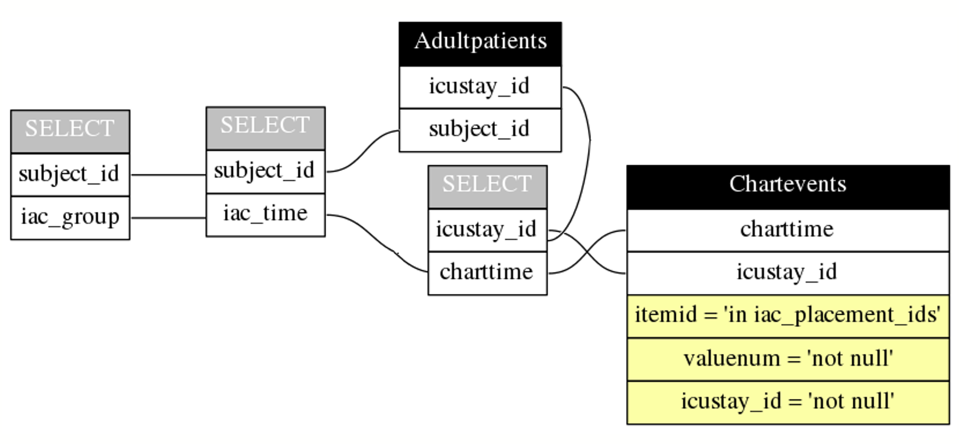}
\caption{MIMIC-III: Divide Cohort}
\label{fig:mimicIII-divideCohort}
\end{figure*}

We combine all criteria to create our final study cohort.
Finally, we divide the cohort into catheterization and control groups, Figure \ref{fig:mimicIII-divideCohort}. This query assigns patients to either the catheterization group (1) or the control group (0) based on the presence of catheterization events in their records.

The data extraction process described in this section provides a comprehensive framework for selecting a study cohort from the MIMIC-III database. By following these steps, researchers can reproduce the data extraction process for studies investigating the relationship between catheterization and mortality in ICU patients. 

\newpage
\section{From Descriptive Analytics to Digital Twins: Advancing Healthcare Information Systems}

\subsection{From Descriptive Analytics to Prescriptive Analytics}


We can conceptualize healthcare information retrieval processes as a pathway from descriptive analytics to diagnostic analytics, predictive analytics, and prescriptive analytics \citep{arjun2019machine, gamal2021standardized}. Descriptive analytics use techniques such as data aggregation, data mining, and intuitive visualizations to provide understanding of historic data. Descriptive analytics provide the answers to questions such as, how many patients were admitted to a hospital last year? How many patients died within 30 days? Or how many patients caught an infection? In other words, descriptive analytics offer intuitive ways to summarize the data via histograms and graphs and show the data distribution properties. Limitations of descriptive analytics are that it keeps limited ability to guide decision because it is based on a snapshot of the past. 

Diagnostic analytics is a form of analytics that examines data to answer the question of why something happened. 
Diagnostic analytics could comprise of correlation techniques that discovers links between clinical variables, treatments, and drugs. Predictive analytics allow us to predict the outcome and likelihood of an event. We may like, for example, to predict the mortality risk of a patient, the length of hospitalization, or the risk for infection. Predictive analytics exploit historic values of the data with the aim to be able to provide useful information about critical events in the future. Predictive analytics are in demand because health care providers would like evidence based ways to predict and avoid adverse events. Importantly predictive analytics enable early intervention which can save patient lives and improve their quality of life. Prescriptive analytics aim to make decisions for optimal outcomes. 

Prescriptive analytics attempt to quantify the effect of future decision to advise on possible outcomes before decisions are made.
Therefore they provide recommendations regarding actions that consider the results of predictive analytics. In other words, prescriptive analytics are important to transition a prediction model to a clinical decision support model.

\subsection{Digital Twins}
\begin{figure*}[h]
\centering
\includegraphics[width=\textwidth]{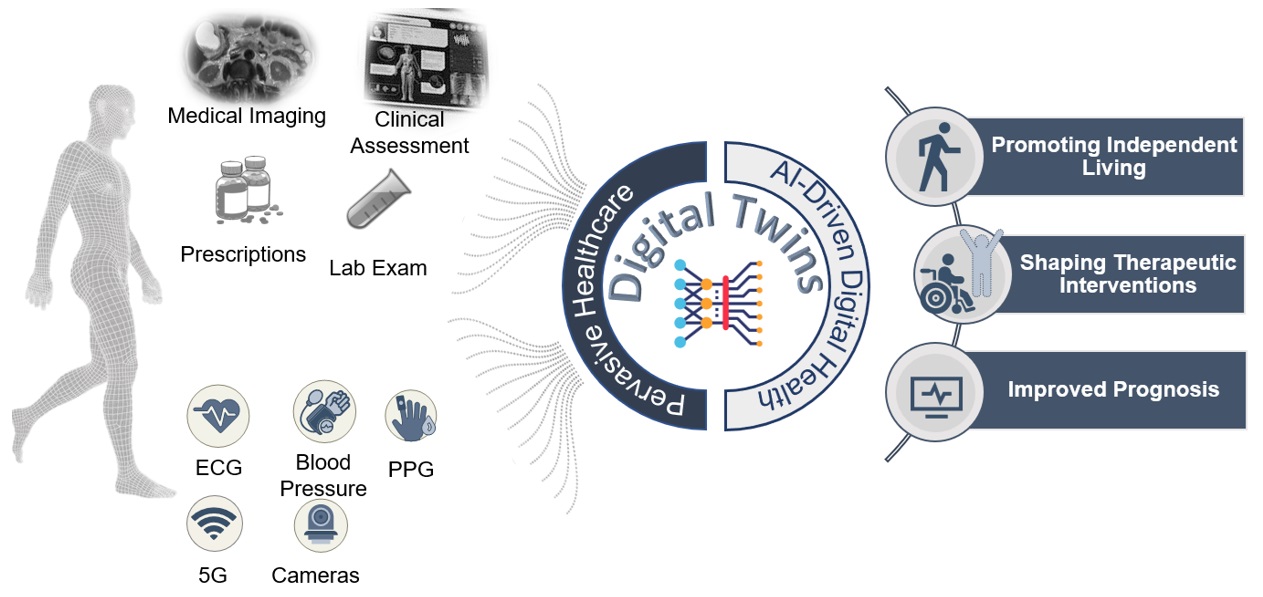}
\caption{Digital Twins in Healthcare}
\label{fig:DigitalTwins}
\end{figure*}

Patients increasingly expect seamless management of their information across their healthcare providers. Currently, traditional healthcare models rely on disconnected systems, multiple sources of information. The new digital healthcare model will transition towards an inherent capability to ensure seamless information exchange across both clinical system and real-time biomarkers of well-being, Figure \ref{fig:DigitalTwins}. In this way, they will update the records of each patient's across of all these systems. This enables data mining and machine learning approaches to successfully applied and advance our knowledge with relation to clinical decision making systems.

‘Digital Twins’ (DTs) is a concept very well known in industrial/manufacturing scale applications and it is used as a virtualisation of physical assets and safety critical processes \citep{rasheed2019digital}. The aim is to simulate decision processes and optimise the design and the underlying operations in a safe and cost-efficient environment \citep{tao2019make}. DTs definitions have been originally formalised by extensive work in large manufacturing companies and NASA, which used the technology to develop digital copies of spacecrafts and monitor its status. In this approach, virtual models are made to duplicate every aspect from design to operation in a safe and cost-efficient way. Originally, DTs were modelled based on three entities, namely, physical entities, virtual models and the connection between physical and virtual entities. This way though could not encompass recent advances in data science and it has been expanded to include DT data and services, which ensure high prognostic fidelity \citep{tao2018digital}. 

It is only recently that it has been realised that DTs have enormous potentials in healthcare and in particular in precision medicine \citep{voigt2021digital, tao2018digital, drummond2023home, sarris2023towards, rasheed2019digital}. The reason behind this shift is that virtual copies of the patients can accelerate the analyses of large amounts of data derived from both clinical practice and sensing modalities in home environments. It has been emphasized that in chronic, multifaceted disorders and diseases such as multiple sclerosis and asthma, DTs can provide unprecedent capabilities of patient centred disease preventive systems along with clinical decision support without increasing the burden on healthcare providers \citep{voigt2021digital, drummond2023home}. For example, telemonitoring devices and sensing technologies in wearables and smart watches can provide digital patient support tools for several diseases such as asthma, multiple sclerosis and diabetes \citep{drummond2023home}. In asthma and diabetes, DTs technology encompasses machine learning models that will provide decision support to impel changes of lifestyle choices and other modifiable risk factors. In multiple sclerosis DTs can simulate therapeutic interventions and provide prognostic tools for disease progression \citep{voigt2021digital}.

\section{Conclusions}

The transition to Electronic Health Records (EHRs) marks a significant milestone in the evolution of healthcare, offering unprecedented opportunities for improving patient care, enhancing clinical decision-making, and advancing biomedical research. By digitizing patient information, EHRs facilitate seamless data integration, enabling healthcare providers to access comprehensive patient histories and make informed decisions. This shift not only improves the efficiency and accuracy of healthcare delivery but also supports the development of personalized medicine and digital twins in healthcare.

The adoption of EHRs in the UK and USA highlights the complexities and challenges involved in transitioning from paper-based systems to digital records. Despite these challenges, the benefits of EHRs are evident in the improved quality of care, enhanced data management, and the potential for large-scale health data analysis. The integration of clinical registries and databases further enriches the healthcare landscape, providing valuable insights into patient outcomes and supporting evidence-based practice.

As we move towards a future where digital twins become an integral part of healthcare, the role of EHRs will continue to expand. These digital records will serve as the foundation for creating accurate and dynamic models of individual patients, enabling real-time monitoring and personalized treatment plans. The ongoing advancements in EHR technology and data analytics will undoubtedly drive innovation in healthcare, ultimately leading to better health outcomes and a more efficient healthcare system.\\

\noindent{\textbf{Acknowledgements}}
Fani Deligianni is supported by funding from EPSRC (EP/W01212X/1) and Academy of Medical Sciences (NGR1/1678). She is also a member of the research team for NIHR (NIHR158303). This chapter builds upon content from the Coursera specialization 'Informed Clinical Decision Making using Deep Learning'.

\newpage
\bibliographystyle{unsrtnat}
\bibliography{references}

\end{document}